\documentclass{article}

\usepackage{arxiv}
\usepackage[utf8]{inputenc} 
\usepackage[T1]{fontenc}    
\usepackage{hyperref}       
\usepackage{url}            
\usepackage{booktabs}       
\usepackage{amsfonts}       
\usepackage{nicefrac}       
\usepackage{microtype}      
\usepackage{graphicx}
\usepackage{doi}
\usepackage{subcaption}
\usepackage{tabularx}

\usepackage{bm}  
\usepackage{amsmath}
\usepackage{natbib}
\usepackage{xurl}

\usepackage{scalerel}
\DeclareMathOperator*{\concat}{\scalerel*{\Vert}{\sum}}
\usepackage{multirow}
\usepackage{enumitem}   

\title{Graph-to-SFILES: Control structure prediction from process topologies using generative artificial intelligence}

\author{ 
    \href{https://orcid.org/0000-0001-7494-9110}{\includegraphics[scale=0.06]{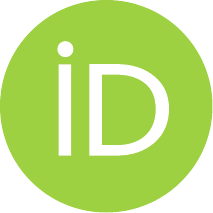}\hspace{1mm}Lukas Schulze Balhorn} \\
	Process Intelligence Research Group\\
	Department of Chemical Engineering\\
	Delft University of Technology\\
	\And
    \hspace{1mm}Kevin Degens \\
	Process Intelligence Research Group\\
	Department of Chemical Engineering\\
	Delft University of Technology\\
    \And
	\href{https://orcid.org/0000-0001-8885-6847}{\includegraphics[scale=0.06]{figures/orcid.pdf}\hspace{1mm}Artur M. Schweidtmann}\thanks{corresponding author} \\
	Process Intelligence Research Group\\
	Department of Chemical Engineering\\
	Delft University of Technology\\
    \texttt{A.Schweidtmann@tudelft.nl} \\
}

\renewcommand{\shorttitle}{Graph-to-SFILES}

\hypersetup{
pdftitle={Graph-to-SFILES},
pdfauthor={L. Schulze Balhorn},
}

\begin{document}
\maketitle

\begin{abstract}    
    Control structure design is an important but tedious step in P\&ID development. 
    Generative artificial intelligence (AI) promises to reduce P\&ID development time by supporting engineers.
    Previous research on generative AI in chemical process design mainly represented processes by sequences. However, graphs offer a promising alternative because of their permutation invariance.
    We propose the Graph-to-SFILES model, a generative AI method to predict control structures from flowsheet topologies.
    The Graph-to-SFILES model takes the flowsheet topology as a graph input and returns a control-extended flowsheet as a sequence in the SFILES 2.0 notation.
    We compare four different graph encoder architectures, one of them being a graph neural network (GNN) proposed in this work.
    The Graph-to-SFILES model achieves a top-5 accuracy of 73.2\% when trained on 10,000 flowsheet topologies. 
    In addition, the proposed GNN performs best among the encoder architectures.
    Compared to a purely sequence-based approach, the Graph-to-SFILES model improves the top-5 accuracy for a relatively small training dataset of 1,000 flowsheets from 0.9\% to 28.4\%. However, the sequence-based approach performs better on a large-scale dataset of 100,000 flowsheets.
    These results highlight the potential of graph-based AI models to accelerate P\&ID development in small-data regimes but their effectiveness on industry relevant case studies still needs to be investigated.
\end{abstract}


\section{Introduction}\label{sec:2023_Graph-to-SFILES_introduction}

Developing piping and instrumentation diagrams (P\&IDs) is currently a manual and tedious task \citep{Toghraei2019_PipingInstrumentationDiagram}. Supporting engineers in P\&ID development with computer algorithms offers potential for cost and time reduction \citep{Dzhusupova2022_PatternRecognitionMethod, Bramsiepe2014_InformationTechnologiesInnovative}.
Time savings can reduce a project's lead time, which is a critical economic success factor, particularly in the fine chemical and pharmaceutical industries \citep{Bramsiepe2014_InformationTechnologiesInnovative}.
In addition, employing computer algorithms for tedious tasks can free up resources for innovation \citep{Dzhusupova2022_PatternRecognitionMethod}.

One essential step in P\&ID development is designing a suitable control structure \citep{Toghraei2019_PipingInstrumentationDiagram}. 
The control structure aims to keep the process within a desired operating window. 
In decentralized control, the control structure consists of controllers that regulate manipulated variables based on measurements to follow the operating window \citep{Morari1980_StudiesSynthesisControl}.
Typically, a suitable control structure is designed from a process flow diagram (PFD) depicting the conceptual design of a process. We refer to a Process Flow Diagram (PFD) that incorporates a decentralized control structure as a control-extended flowsheet (CEF).
Control structure design involves, among other tasks, analyzing the process degrees of freedom (i.e., the number of controllable variables) and defining control tasks, such as set point tracking or disturbance rejection, based on the operating window \citep{Morari1980_StudiesSynthesisControl,Luyben1997_PlantwideControlDesign,Ng1998_SynthesisControlStructures}.
Well-studied methods like dynamic simulations \citep{Seborg2016_ProcessDynamicsControl}, relative gain array \citep{VanDeWal2001_ReviewMethodsInput/output}, and further methods assist engineers.

Previous research on automatic control structure design has led to several promising approaches. The first approaches include knowledge-based expert systems \citep{Williamson1989_ComputerAidedProcess,Song1990_Intellite3Knowledge} and heuristic design procedures \citep{Luyben1997_PlantwideControlDesign}. 
Despite these achievements, expert systems are generally expensive to develop and once set up, expert systems tend to be labor-intensive and expensive to maintain and extend \citep{Venkatasubramanian2019_PromiseArtificialIntelligence}.
Recent advancements in machine learning, specifically in deep learning-based artificial intelligence (AI) have demonstrated remarkable improvements over traditional expert systems in several domains such as natural language processing (NLP) \citep{Popel2020_TransformingMachineTranslation,Brown2020_LanguageModelsAre}. 
Also in the natural sciences and engineering, deep learning has outperformed rule-based approaches. For example in organic chemistry, a transformer-based model can accurately predict reaction outcomes \citep{Schwaller2019_MolecularTransformerModel,Tu2023_PermutationInvariantGraph}. Here, rule-based models used molecular templates selected by experts.
Chemical engineering is another field where AI holds great potential to assist engineers with design tasks \citep{Schweidtmann2024_GenerativeArtificialIntelligence}. 
This motivates the investigation of AI-based control structure prediction in this research work. 
Recently, we proposed a generative AI model capable of extending a given PFD with a control structure \citep{Hirtreiter2023_AutomaticGenerationControl}. \citet{Hirtreiter2023_AutomaticGenerationControl} formulate the control structure design as a translation task, i.e., a sequence-to-sequence task, converting a PFD without a control structure into a CEF. The PFD and CEF are represented using the SFILES 2.0 string format \citep{Vogel2023_Sfiles2.0Extended}. When trained on a synthetic dataset consisting of 100,000 PFD-CEF pairs, the generative AI model was able to predict the correct CEF for a given PFD with an accuracy of 48.6\%. 

An essential design choice for an AI model is the process information representation. The two most common process information representations are (i) sequences and (ii) graphs. 
First, generative AI models for chemical process development so far mainly represent processes as sequences, i.e., as strings \citep{Nabil2019_MachineLearningBased,Nabil2023_DataDrivenStructural, Vogel2023_LearningFlowsheetsGenerative,Balhorn2024_AutocorrectionChemicalProcess,Hirtreiter2023_AutomaticGenerationControl,Oeing2022_UsingArtificialIntelligence,Azevedoa2024_RepresentationLearningFlowsheets,Li2024_LearningHybridExtraction}. The elements of the sequence can represent equipment and the order of the elements defines their connectivity \citep{dAnterroches2005_ProcessFlowsheetGeneration,Vogel2023_Sfiles2.0Extended,Mann2024_EsfilesIntelligentProcess}.
Second, a process can be represented as a graph. A graph is a mathematical structure consisting of a set of nodes and a set of edges that describe relations between nodes. In the context of chemical processes, nodes can represent equipment and edges can represent pipes or signal lines between the equipment. The potential of graphs in AI for process development has already been shown for P\&ID verification using a graph neural network (GNN) \citep{Oeing2022_UsingArtificialIntelligence,Oeing2023_GraphLearningMachine} and process synthesis using reinforcement learning \citep{Stops2023_FlowsheetGenerationHierarchical}.
The choice of information representation also influences what kind of generative models can be used. Graph-based approaches for generative AI commonly rely on Variational Graph Autoencoders (VGAEs) \citep{Kipf2016_VariationalGraphAuto}, Graph Recurrent Neural Networks (GraphRNNs) \citep{You2018_GraphrnnGeneratingRealistic}, or Graph Generative Adversarial Networks (Graph GANs) \citep{DeCao2018_MolganImplicitGenerative}, while sequence-based approaches rely on Recurrent Neural Networks (RNNs) \citep{Hochreiter1997_LongShortTerm} or the transformer \citep{Vaswani2017_AttentionIsAll}.
It remains an open research question which form of information representation is most effective for chemical processes.

Graph-to-sequence models promise to combine the advantages of both sequence and graph representation.
Graph-to-sequence models consist of two main components: (i) A graph encoder that learns a numerical representation of the input graph and (ii) a sequence decoder that, given the numerical representation, generates a sequence.
Graph-to-sequence models were first proposed in NLP \citep{Beck2018_GraphSequenceLearning, Song2018_GraphSequenceModel, Xu2018_Graph2seqGraphSequence}. 
Also in the field of chemistry, graph-to-sequence-based models have been used for various tasks such as predicting the outcome of a reaction \citep{Mao2021_MolecularGraphEnhanced,Tu2023_PermutationInvariantGraph}, developing new molecules with favorable properties \citep{Wang2023_MetaLearningLow,Dollar2023_Efficient3dMolecular}, or synthetic polymer design \citep{Vogel2023_GraphStringVariational}. 

We hypothesize that, in the context of flowsheets, a graph encoder is more data efficient than a sequence encoder for two reasons:
First, the sequence encoder captures the flowsheet topology explicitly.
In the message passing of the graph encoder, information is only exchanged between nodes (here: flowsheet component) that are connected via an edge (here: pipe or signal line).  
The sequence encoder treats the input as a fully connected graph. In the sequence attention mechanism, every token (here: flowsheet component, among others) attends to every other token. Sequence-based approaches need to learn the grammar of the sequence (here: SFILES 2.0 notation) to infer the flowsheet topology.
Second, the graph encoder is permutation invariant whereas the sequence decoder is permutation sensitive. A function is permutation invariant if permutations to the input do not affect the output.
Permutation invariance is desirable for creating a unique graph embedding \citep{Tu2023_PermutationInvariantGraph}.
For sequence representations of graphs such as SMILES or SFILES, the model has to “re-learn” or “re-discover” that different sequences share the same underlying graph, which can demand more data or complex augmentations.
Another advantage of the graph input is that it is more intuitive to include detailed process data, such as operating points, in a graph \citep{Stops2023_FlowsheetGenerationHierarchical} than in a sequence.
In contrast, generating a sequence representation such as SMILES or SFILES is often reported to be computationally more efficient than generating graphs, particularly in machine learning applications \citep{Dollar2023_Efficient3dMolecular}. A possible explanation for this efficiency is that (i) the structured nature of sequences (here: the grammar of SFILES) simplifies autoregressive decoding in transformer-based models, and (ii) the search space is smaller when generating a linear representation (selecting a token) compared to generating a graph with arbitrary connectivity (selecting a source-target pair and a node type).
Therefore, a beneficial inductive bias is to use a sequence decoder and represent the output (the generated CEF) as a sequence.

Previous studies have shown that graph-to-sequence models can improve accuracy compared to purely graph-based \citep{Dollar2023_Efficient3dMolecular} or purely sequence-based models \citep{Tu2023_PermutationInvariantGraph}.
To the best of our knowledge, no graph-based, generative AI method for process development exists yet.

We propose a Graph-to-SFILES model for predicting control structures to aid P\&ID development. Thereby, we use permutation invariant graphs to represent PFDs as the model input. 
To account for the variety of graph encoders available in literature, we compare three state-of-the-art GNN architectures: (i) The Graph Attention Transformer (GATv2) \citep{Velickovic2017_GraphAttentionNetworks, Brody2021_HowAttentiveAre}, (ii) the Graph Convolution Transformer (GraphConv) \citep{Shi2020_MaskedLabelPrediction}, (iii) the Graph Transformer \citep{Dwivedi2020_GeneralizationTransformerNetworks}.
These architectures are selected based on their relevance in the literature and on their effectiveness in learning graph representations through attention-based message passing \citep{Min2022_TransformerGraphsOverview,Khemani2024_ReviewGraphNeural}. 
In addition, we propose a GNN architecture for the graph encoder that integrates key aspects of the selected GNNs. 
The representation of the output CEF is chosen as SFILES for efficient CEF generation.
Inspired by the success in other domains, we hypothesize that for chemical processes a graph-to-sequence model yields better performance than a purely sequence-based model.

The remainder of this article is structured as follows: Section \ref{sec:2023_Graph-to-SFILES_inf_repr} introduces the information representations used in this work: Process graphs and SFILES 2.0. In Section \ref{sec:2023_Graph-to-SFILES_method}, we describe the Graph-to-SFILES methodology for control structure prediction. Afterward, we explain the data used to train the Graph-to-SFILES model (Section \ref{sec:2023_Graph-to-SFILES_data}). In Section \ref{sec:2023_Graph-to-SFILES_results}, we discuss the model's results and demonstrate its performance in an illustrative case study.

\section{Information representation of flowsheets}\label{sec:2023_Graph-to-SFILES_inf_repr}

\begin{figure} [h]
    \centering
    \begin{subfigure}[b]{0.45\textwidth}
        \centering
        \includegraphics[width=\textwidth]{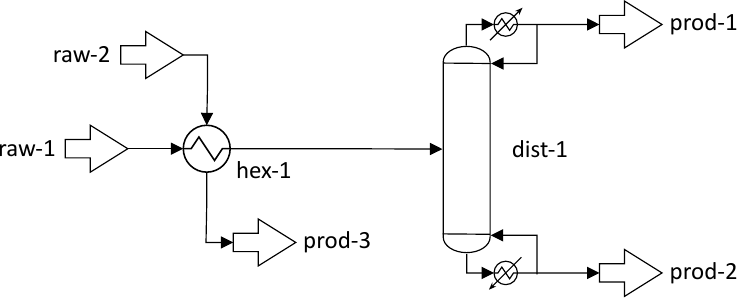}
        \caption{PFD example.}
        \label{fig:2023_Graph-to-SFILES_example_PFD}
    \end{subfigure}
    \hfill
    \begin{subfigure}[b]{0.45\textwidth}
        \centering
        \includegraphics[width=\textwidth]{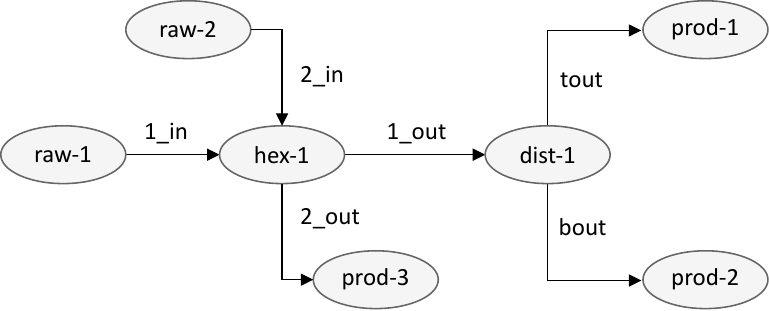}
        \caption{PFD example represented as a graph.}
        \label{fig:2023_Graph-to-SFILES_example_graph}    
    \end{subfigure}
    \caption{Graph representation of flowsheets. The corresponding SFILES 2.0 notation is: \texttt{(raw)(hex)\{1\}(dist)[\{bout\}(prod)]\{tout\}(prod)n|(raw)(hex)\{1\}(prod)}.}
    \label{fig:2023_Graph-to-SFILES_example}
\end{figure}

To leverage machine learning and learn from chemical processes and their data, we need to represent chemical processes in a machine-readable format. We represent processes as graphs for the input to our models (Section \ref{sec:2023_Graph-to-SFILES_inf_repr_process_graphs}) and as SFILES for the output of our models (Section \ref{sec:2023_Graph-to-SFILES_inf_repr_sfiles}). The distillation column in Figure \ref{fig:2023_Graph-to-SFILES_example_PFD} serves as an example to illustrate the information representations.

\subsection{Process graphs}\label{sec:2023_Graph-to-SFILES_inf_repr_process_graphs}

Chemical processes can be viewed as complex networks of unit operations connected through streams. An intuitive choice for the information representation is therefore a graph. We make use of the graph format defined by \citet{Vogel2023_Sfiles2.0Extended} for PFDs. Here, nodes represent unit operations, control units, raw materials, and products, while directed edges represent streams and signals.
Control units that refer to a stream, such as a temperature indicator, are placed in the stream by splitting the stream into two.
In addition, edges have attributes to store further stream information: If multiple outlet streams exist, we define the outlet location as top outlet (\texttt{tout}) or bottom outlet (\texttt{bout}). For heat exchangers, we identify the incoming and outgoing streams that belong to the same mass train using the edge attribute \texttt{StreamID\textunderscore\{in/out\}}.
Figure \ref{fig:2023_Graph-to-SFILES_example_graph} shows an example process represented as a graph.

\subsection{SFILES}\label{sec:2023_Graph-to-SFILES_inf_repr_sfiles}

An alternative information representation to graphs is the Simplified Flowsheet Input Line Entry System (SFILES). The SFILES was developed by \citet{dAnterroches2005_ProcessFlowsheetGeneration} and is inspired by the Simplified Molecular Input Line Entry system (SMILES) \citep{Weininger1989_Smiles.2.Algorithm} used to encode molecular graphs to strings. The SFILES encodes the topology of a process as a sequence of tokens, such as \texttt{(raw)} for the raw material or \texttt{(r)} for a reactor. The order of the tokens is determined by following a path from the raw materials to the products along the streams.
In SFILES, square brackets indicate a side stream that diverges from the main process stream. At the place where the side stream diverges the brackets are opened, and the tokens of that side stream are added. When a side stream rejoins the main stream or ends the brackets are closed and the SFILES notation continues from where the stream diverged.

Recently \citet{Vogel2023_Sfiles2.0Extended} introduced SFILES 2.0, which extends the original SFILES notation with control structures and topology details.
Controllers are incorporated using the controller token \texttt{(C)} followed by a tag with the letter code (e.g., \texttt{\{PC\}}). The position of the controller token is based on its location in the flowsheet. The signal line is encoded similarly to recycles and differentiated by an underscore (\texttt{\_\#}, \texttt{<\_\#}). 
Topology details such as a stream leaving a column at the top are incorporated as tags in curly brackets (e.g., \texttt{\{tout\}}). By using an extra tag before a unit, the SFILES 2.0 encodes where the stream enters a unit, and similarly, by adding a tag in front of the stream leaving a unit it is encoded where the stream leaves.
Independent material streams are appended to the SFILES, starting with an \texttt{n|}.
Note that there can be multiple SFILES for the same flowsheet but there exists a unique, or canonical, way of writing them, defined by a graph invariant.
The graph invariant provides a unique rank for each node that determines the order of the SFILES sequence. The graph invariant is obtained in two steps.
First, the Morgan algorithm \citep{Morgan1965_GenerationUniqueMachine} determines a node ranking.
Second, a set of rules determines the order of the tokens in case the Morgan algorithm does not yield a unique rank for every node. These rules define an order based on (i) the node type (control, outlet, inlet, other), (ii) the number of successors in a depth-first search tree of the graph, (iii) the alphabetic order of the node name, and (iv) the node number (e.g., if the flowsheet contains three heat exchangers, they will be numbered from one to three).
These rules are described in \citep{Vogel2023_Sfiles2.0Extended} and made accessible via a Python package\footnote{\url{https://github.com/process-intelligence-research/SFILES2}}.
To illustrate the notation, we provide the SFILES 2.0 of the distillation column from Figure \ref{fig:2023_Graph-to-SFILES_example} in the figure caption.

\section{Graph-to-SFILES methodology}\label{sec:2023_Graph-to-SFILES_method}

\begin{figure}[h]
    \centering
    \includegraphics[width=.7\textwidth]{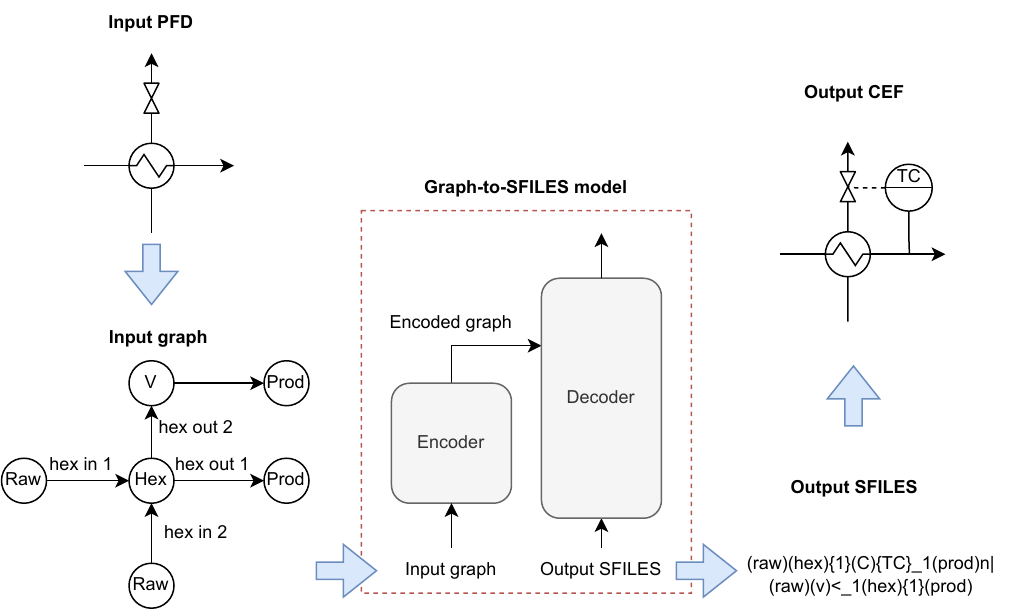}
    \caption{Overview of Graph-to-SFILES methodology. The PFD is represented as a graph with node and edge features. The Graph-to-SFILES model translates the graph to an SFILES that describes a potential CEF for the given PFD. Note that the control structure shown here is one of multiple possible control options.}
    \label{fig:2023_Graph-to-SFILES_methodology}
\end{figure}

\begin{figure}[h]
    \centering
    \includegraphics[scale=0.7]{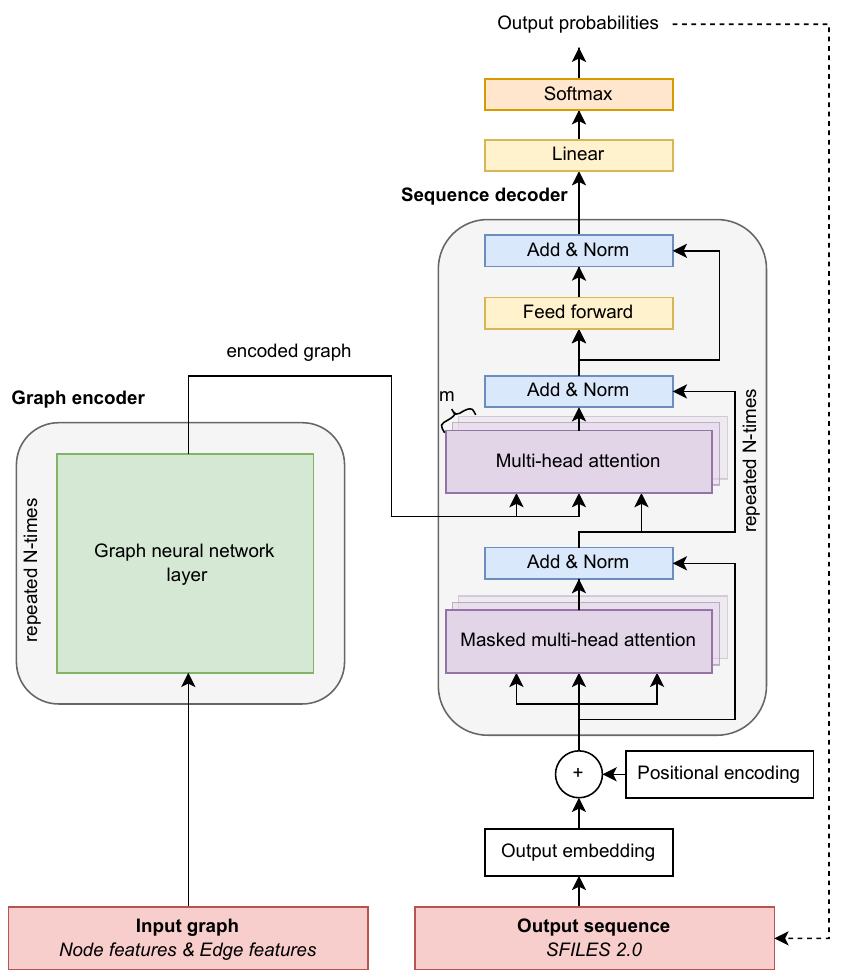}
    \caption{Detailed Graph-to-SFILES model consisting of a graph encoder and a sequence decoder.}
    \label{fig:2023_Graph-to-SFILES_model}
\end{figure}

We propose the Graph-to-SFILES model to predict the control structure of chemical processes. Figure \ref{fig:2023_Graph-to-SFILES_methodology} provides an overview of the model.
We start with a PFD that is represented by a graph as described in Section \ref{sec:2023_Graph-to-SFILES_inf_repr_process_graphs}. Given the graph, the Graph-to-SFILES model generates an SFILES that represents the CEF (Section \ref{sec:2023_Graph-to-SFILES_inf_repr_sfiles}).
The Graph-to-SFILES model is inspired by the Graph2SMILES model for molecules \citep{Tu2023_PermutationInvariantGraph} and is based on a transformer architecture, as shown in Figure \ref{fig:2023_Graph-to-SFILES_model}.
The model consists of two main components: (i) A graph encoder to encode a PFD given as a process graph and (ii) a sequence decoder that, given the encoded input, generates the CEF in the SFILES 2.0 format. 
In Section \ref{sec:2023_Graph-to-SFILES_method_encoder}, we describe the graph encoder. 
The graph encoder aims to convert the input process graph into a continuous latent-space representation. The latent-space representation is essential for the sequence decoder to generate a suitable control structure. 
The sequence decoder and the corresponding decoding strategy are described in Section \ref{sec:2023_Graph-to-SFILES_method_decoder} and Section \ref{sec:2023_Graph-to-SFILES_method_decoding_strategy} respectively.
Then, we provide details on the training process in Section \ref{sec:2023_Graph-to-SFILES_method_training}.

\subsection{Graph encoder}\label{sec:2023_Graph-to-SFILES_method_encoder}

We implement the graph encoder as a GNN.
GNNs are a type of neural network designed to operate on graph-structured data. 
The GNN iteratively aggregates information from neighboring nodes (and edges, depending on the architecture) to update the node (and edge) features. We define a sequence of GNN layers, where each layer updates the node features based on their direct neighboring nodes and edges.

A GNN layer generally consists of two basic steps, a message passing step and an update step. During the message passing step, each node sends a message, usually a linear transformation of its feature vector, to its neighboring nodes.
The message passing is followed by an update step in which the incoming messages are merged into a single vector. This step uses a permutation invariant function, e.g., $sum$, $mean$, $min$, or $max$. 
Information is shared across the graph by using multiple iterations of these message-passing and update steps. The linear transformations using learnable parameters in the message passing and update steps allow the learning of complex relationships between the nodes and edges, making their feature vectors more valuable for downstream tasks. In the Graph-to-SFILES model, the individual component nodes get information about their neighboring components and streams.

\begin{figure}[h]
    \centering
    \begin{subfigure}[b]{0.25\textwidth}
        \includegraphics[width=0.95\textwidth]{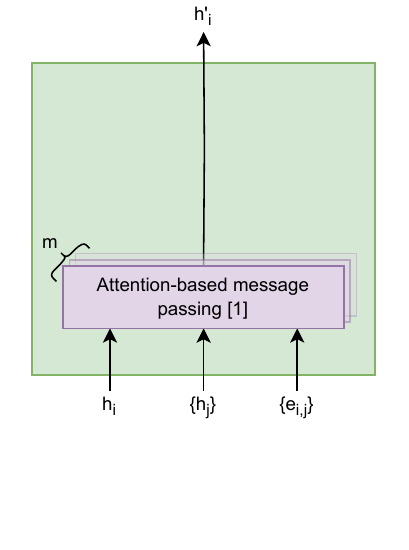}
        \caption{GATv2 \citep{Velickovic2017_GraphAttentionNetworks,Brody2021_HowAttentiveAre}}
        \label{fig:Graph-to-SFILES-GATv2}
    \end{subfigure}
    \hfill
    \begin{subfigure}[b]{0.25\textwidth}
        \includegraphics[width=0.95\textwidth]{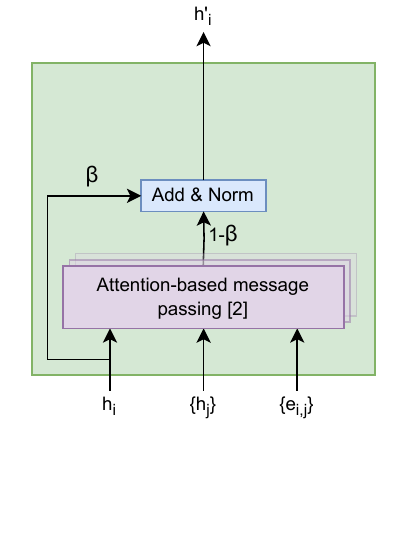}
        \caption{GraphConv \citep{Shi2020_MaskedLabelPrediction}}
        \label{fig:Graph-to-SFILES-GraphConv}
    \end{subfigure}
    \hfill
    \begin{subfigure}[b]{0.25\textwidth}
        \includegraphics[width=\textwidth]{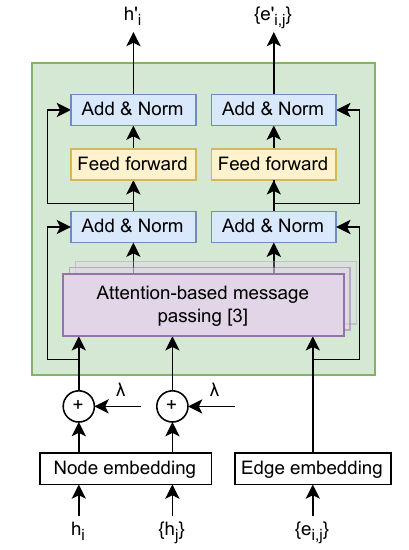}
        \caption{Graph Transformer \citep{Dwivedi2020_GeneralizationTransformerNetworks}}
        \label{fig:Graph-to-SFILES-GraphTransformer}
    \end{subfigure}
    \hfill
    \begin{subfigure}[b]{0.25\textwidth}
        \includegraphics[width=0.9\textwidth]{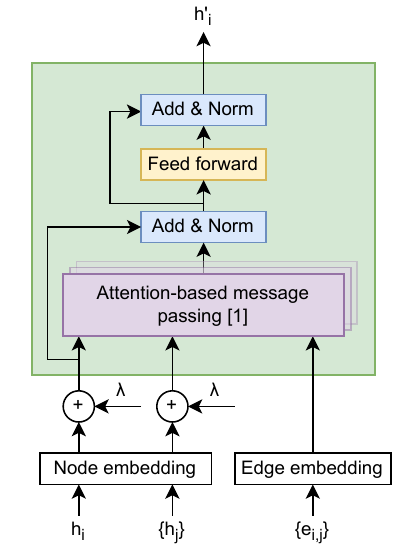}
        \caption{Combined model}
        \label{fig:Graph-to-SFILES-hybrid}
    \end{subfigure}
    \caption{GNN layers as a drop-in replacement in the graph encoder. Each GNN layer takes the node features $\bm{h}$ and the edge features $e$ as input and returns the updated node feature $\bm{h}^{'}$ (and updated edge features $\bm{e}^{'}$ for the Graph Transformer). The attention-based message passing is calculated for $m$ heads in parallel.}
    \label{fig:Graph-to-SFILES-GNN-layers}
\end{figure}

We propose and compare four different GNN architectures for the generation of CEFs from PFDs, as shown in Figure \ref{fig:Graph-to-SFILES-GNN-layers}: (i) The Graph Attention Transformer (GATv2) \citep{Velickovic2017_GraphAttentionNetworks, Brody2021_HowAttentiveAre}, (ii) the Graph Convolution Transformer (GraphConv) \citep{Shi2020_MaskedLabelPrediction}, (iii) the Graph Transformer \citep{Dwivedi2020_GeneralizationTransformerNetworks}, and (iv) a new GNN that we propose in this study, the Combined model.
The GAT model \citep{Velickovic2017_GraphAttentionNetworks} was the first model to introduce the attention mechanism for GNNs and outperforms previous graph convolution architectures. 
Similarly to the GAT model, \citet{Shi2020_MaskedLabelPrediction} proposed an attention-based GNN. They report an improved performance compared to the GAT model by introducing a different attention mechanism and including gated residual connections.
In addition, the Graph Transformer \citep{Dwivedi2020_GeneralizationTransformerNetworks} generalizes the transformer architecture from sequences to graphs. As the Graph Transformer is closely aligned with the transformer encoder architecture, it is expected to integrate well with the overall Graph-to-SFILES transformer approach. \citet{Dwivedi2020_GeneralizationTransformerNetworks} state, that the Graph Transformer can be integrated as a black box in various applications due to its simplicity and generality.
All architectures utilize the attention mechanism to facilitate the message passing, where attention is the weight of the message between node $i$ and node $j$, denoted as $\alpha_{i,j}$.

Here, we describe the main mathematical details of the Graph-to-SFILES encoder. The detailed mathematical expressions of the Combined model are described in the Appendix \ref{sec:2023_Graph-to-SFILES_appendix_model}.
Firstly, we implement the GATv2 as proposed by \citet{Velickovic2017_GraphAttentionNetworks} and modified by \citet{Brody2021_HowAttentiveAre}
(Figure \ref{fig:Graph-to-SFILES-GATv2}). In the attention-based message passing block the attention $\alpha_{i,j}\in\mathbb{R}$ is calculated using

\begin{equation}
\alpha_{i,j} = \text{softmax}\left(\bm{a}^{\top} \text{LeakyReLU}\left(\left[\bm{W}_{q_{\rm enc}}\bm{h}_{i}\Vert\bm{W}_{k_{\rm enc}}\bm{h}_{j}\Vert\bm{W}_{e_{\rm enc}}\bm{e}_{i,j}\right]\right)\right) \nonumber
\end{equation}

The inputs consisting of node features $\bm{h}\in\mathbb{R}^{d_{\rm enc}}$ and edge features $\bm{e}\in\mathbb{R}^{d_{\rm enc}}$ are multiplied by weight matrices $\bm{W}\in\mathbb{R}^{d_{\rm enc}\times d_{\rm enc}}$ and then concatenated. After applying the nonlinear function LeakyReLU, the output is multiplied by another weight matrix $\bm{a}\in\mathbb{R}^{3d_{\rm enc}}$ and normalized using the softmax function. Here, $d_{\rm enc}$ is the internal encoder dimension. Note that the first encoder layer uses $\bm{W}_{q_{\rm enc}},\bm{W}_{k_{\rm enc}}\in\mathbb{R}^{d_{node}\times d_{\rm enc}}$ and $\bm{W}_{e_{\rm enc}}\in\mathbb{R}^{d_{\rm edge}\times d_{\rm enc}}$ to map the node $d_{node}$ and edge $d_{\rm edge}$ dimension to the encoder dimension.
After calculating the attention between the nodes, we determine the updated node features $\bm{h}^{'}$:

\begin{equation}
\bm{h}^{'}_i =  \text{ReLU} \left( \sum_{j \in \mathcal{N}_{i}}{\alpha_{i,j} \bm{W}_{v_{\rm enc}}  \bm{h}_j}  \right). \nonumber
\end{equation}

For each node $i$, the message from node $j$ in $\mathcal{N}_{i}$ is given as $\alpha_{i,j}\bm{W}_{v_{\rm enc}}\bm{h}_j$ with $\mathcal{N}_{i}$ being the direct neighbors of node $i$ including itself. In the update step, the messages are summed up, followed by a nonlinearity (ReLU). 

Secondly, we implement the GraphConv (Figure \ref{fig:Graph-to-SFILES-GraphConv}). The GraphConv utilizes scaled dot-product attention, where the attention $\alpha_{i,j}$ is given by

\begin{equation}
\alpha_{i,j} = \text{softmax} \left( \frac{(\bm{W}_{q_{\rm enc}}\bm{h}_{i})  (\bm{W}_{k_{\rm enc}}\bm{h}_{j} +\bm{W}_{e_{\rm enc}}\bm{e}_{i,j} )^\top}{\sqrt{d_{\rm enc}}}\right) .\nonumber
\end{equation}

In addition, the GraphConv makes use of a gated residual connection. The gated residual connection should prevent the graph encoder from over-smoothing by injecting the residual node feature $\bm{h}_i$ directly in the updated node feature $\bm{h}'_i$. Here, a weight ($\beta=f(\bm{h}_{i},\bm{h}_{j},\bm{e}_{i,j}) \in [0,1]$) is calculated which determines the importance between the residual and the message passing.
Then, the outcome of the gated residual connection is normalized using layer normalization and passed through a final nonlinearity (ReLU). This results in an updated node feature ($\bm{h}'_{i}$). 

Thirdly, we implement the Graph Transformer (Figure \ref{fig:Graph-to-SFILES-GraphTransformer}). The attention used in their model is defined as

\begin{equation}
\alpha_{i,j} = \text{softmax}\left( \frac{(\bm{W}_{q_{\rm enc}}\bm{h}_{i}) (\bm{W}_{k_{\rm enc}}\bm{h}_{j})^\top }{\sqrt{d_{\rm enc}}}\cdot \bm{W}_{e_{\rm enc}}\bm{e}_{i,j} \right). \nonumber
\end{equation}

The unique characteristic of the Graph Transformer is the positional encoding. While the message passing step passes messages only from direct neighbors of a node, the positional encoding encodes the complete graph structure in each node feature. \citet{Dwivedi2020_GeneralizationTransformerNetworks} use eigenvectors of the graph's Laplacian matrix to create a positional encoding $\bm{\lambda}\in\mathbb{R}^{d_{\rm enc}}$. The Laplacian positional encoding is added to the node features in the first GNN layer that also includes a linear transformation for node ($d_{\rm node}\rightarrow d_{\rm enc}$) and edge embedding ($d_{\rm edge}\rightarrow d_{\rm enc}$).
Furthermore, the Graph Transformer includes two residual connections after the attention-based message passing with a feed-forward neural network (FFN) between the residual connections.

Fourthly, we propose and implement the Combined model, which combines the dynamic attention mechanism from GATv2 with the positional encoding and residual connections from the Graph Transformer architecture, as shown in Figure \ref{fig:Graph-to-SFILES-hybrid}.

Similar to the attention used in the work of \citet{Vaswani2017_AttentionIsAll}, multi-head attention is used for all graph encoders to capture multiple patterns during the learning. Instead of calculating only one attention weight, $m$ attention weights are calculated in parallel where each attention calculation is called a head. The outcomes of the $m$ heads are concatenated for the remaining operations.
Finally, the updated node features from the last GNN layer, representing the encoded process graph, are passed to the sequence decoder.

\subsection{Sequence decoder}\label{sec:2023_Graph-to-SFILES_method_decoder}

The decoder generates the CEF in the SFILES 2.0 notation, given the PFD as an encoded process graph. We follow the original implementation of the transformer decoder by \citet{Vaswani2017_AttentionIsAll}, as shown in Figure \ref{fig:2023_Graph-to-SFILES_model}.

The decoder generates an output sequence iteratively. The first token of the output sequence is always a start-of-sequence token \texttt{(SOS)}. Then, every forward pass of the decoder adds a new token to the output sequence until the new token is an end-of-sequence token \texttt{(EOS)}. This generation technique is called autoregression, meaning the model generates its output step-by-step, based on previously generated tokens and the input sequence. Autoregression allows the model to create novel and coherent sequences.

Before the output tokens are processed by the decoder, they are transformed into continuous vector representations, or embeddings, of dimension $d_{dec}$ using a linear transformation.
The embedding is complemented by positional encoding. The decoder does not contain a function that keeps track of the sequence order, like recurrence or convolutions. To address this, the Graph-to-SFILES model uses pairs of sinusoidal functions \citep{Vaswani2017_AttentionIsAll}. A positional encoding vector is calculated for every input token and added to the embedding, allowing the model to differentiate identical tokens at different positions. 
The resulting embedding is fed into the decoder.

The decoder contains multiple layers and each layer contains blocks of attention mechanisms and FFNs.
First, the decoder analyzes the embedded output $\bm{x}\in\mathbb{R}^{d_{dec}}$ with the scaled dot-product attention mechanism as described by

\begin{equation}
\text{Attention}(\bm{x}) = \text{softmax}\left(
\frac{(\bm{W}_{q_{dec}}\bm{x})(\bm{W}_{k_{dec}}\bm{x})^\top}{\sqrt{d_{dec}}}\right)\bm{W}_{v_{dec}}\bm{x}. \nonumber
\end{equation}

Similar to the attention in the graph encoder (Section \ref{sec:2023_Graph-to-SFILES_method_encoder}), attention describes the importance of one token for the context of another token. 
The attention mechanism is enhanced by using $m$ attention heads that calculate the attention independently. The resulting outputs of each attention head are concatenated as introduced in Section \ref{sec:2023_Graph-to-SFILES_method_encoder}.
Second, we introduce the encoded process graph to the decoder. The encoded graph is multiplied by the query matrix $\bm{W}_{q_{dec}}$ and key matrix $\bm{W}_{k_{dec}}$ in a further attention block. The output of the previous attention block is multiplied by the value matrix $\bm{W}_{v_{dec}}$.
Third, an FFN is applied to transform the output of the self-attention mechanism. Each block has a residual connection where the block's output and the residual are summed and normalized using $LayerNorm$.
Finally, the sequence decoder's output is mapped to a probability for the next token to be added to the output sequence.

\subsection{Decoding strategy}\label{sec:2023_Graph-to-SFILES_method_decoding_strategy}

The decoding strategy defines how the next token of the output sequence is selected. The computationally cheapest strategy is greedy search. At every prediction step of the decoder, greedy search selects the token with the highest probability. However, selecting the token with the highest probability likely does not lead to the sequence with the highest joint probability. In addition, the coherence of the output sequence can be suboptimal, with subsequences that are not related to each other. 
In contrast to greedy search, calculating all possible sequences and comparing their joint probability is intractable.

To overcome the aforementioned limitations, we use beam search as the decoding strategy.
At the first decoding step, beam search keeps track of the $k$ tokens with the highest probability, where each token represents a beam.
In the next decoding step, the model returns a probability distribution for the next token of each beam. Using this distribution, the joined probability of extending the beams is calculated. Then, the $k$ combinations yielding the highest joined probabilities are selected and used as beams for the next decoding step. This process is repeated until all beams have reached the end-of-sequence token (<EOS>) or a maximum number of tokens. Then, the model returns the beams in order of highest joint probability, yielding top-k SFILES predictions. 

\subsection{Training}\label{sec:2023_Graph-to-SFILES_method_training}

The objective function for training the Graph-to-SFILES model is the cross-entropy loss that compares the probability of the predicted next token with the ground truth token.
During the training, we use teacher-forcing. Teacher-forcing is a training strategy for supervised learning, c.f., if the correct output sequence is known. At each decoding step, the model output is corrected to force the model to predict the correct output sequence for the given input sequence.
The Graph-to-SFILES model is implemented in Python using the PyTorch and PyTorch Geometric libraries.

\section{Data}\label{sec:2023_Graph-to-SFILES_data}

We train the proposed Graph-to-SFILES model on a dataset containing 100,000 pairs of PFDs and CEFs which were synthetically generated from a collection of process patterns from literature. Every pattern contains a decentralized control structure. The patterns are combined following engineering heuristics to generate complete chemical process schematics. Each process starts with up to three feed streams which may be preprocessed by a temperature change, pressure change, or mixing. Then, either thermal separation or reaction is selected. This step is either repeated or the process is terminated and the products are conditioned. 
For further details of the dataset, we refer the interested reader to \citep{Hirtreiter2023_AutomaticGenerationControl}.
The resulting flowsheets primarily represent oil and gas processes, as these are commonly simulated in Aspen Plus. It includes unit operations such as gas-phase reactors, heat exchangers, compressors, pumps, and distillation columns, along with their associated control structures.
Additionally, often multiple valid control options exist. This is reflected in our dataset. For instance, we extend the PFD topology of a distillation column with seven different control structures. 
An overview of the process patterns used to generate the synthetic data can be found in the Supplementary Information of \citet{Hirtreiter2023_AutomaticGenerationControl}.
We report dataset statistics in Table \ref{tab:Graph-to-SFILES_dataset_properties}
While we aim to develop a dataset of flowsheets that resembles oil and gas processes by mimicking their structure and scale, our flowsheets are simplified and do not fully represent an actual industrial process. The diversity and complexity of equipment and control structures in industrial processes are significantly higher. Additionally, since we sample patterns based on a given probability distribution, there is a possibility that incorrect combinations of decentralized control structures may be selected. 
Furthermore, the current dataset focuses on oil and gas processes. Extending the methodology to other fields, such as bioprocessing, would require developing a new dataset that includes relevant unit operations (e.g., fermenters, liquid-liquid membrane separators) and their corresponding control strategies. However, this is a non-trivial task and comes with a significant workload.
Creating a dataset for a different domain would involve: (i) Collecting representative process topologies from sources such as process simulators, literature, patents, or industrial case studies, (ii) defining appropriate control structures for the new domain, as control strategies may differ significantly from those in oil and gas, (iii) extending the SFILES 2.0 notation to include any missing unit operations, and (iv) curating and verifying the dataset to ensure it captures realistic process control principles.

There are two versions of the dataset. First, the controllers are removed from the flowsheet to create the PFD from the CEF. 
Secondly, the valves were removed in addition to make the schematics more in line with a flowsheet found in simulation software like Aspen. The usage of this dataset will be indicated as without valves. 

\begin{table}[h]
    \centering
    \caption{Dataset properties.}
    \label{tab:graph_stats}
    \begin{tabular}{l c}
        \toprule
        \textbf{Statistic} & \textbf{Value} \\
        \midrule
        Training samples & 1000 / 10000 / 100000 \\
        Validation samples & 1000 \\
        Test samples & 1000 \\
        Average number of nodes & $52 \pm 20$ \\
        Average number of edges & $61 \pm 23$ \\
        Average out-degree & $1.2$ \\
        Vocabulary size & 113 \\
        \bottomrule
    \end{tabular}
	\label{tab:Graph-to-SFILES_dataset_properties}
\end{table}

Pre-processing of the input data is needed to turn the SFILES into the graph format from Section \ref{sec:2023_Graph-to-SFILES_inf_repr_process_graphs}. Firstly, we utilize the open-source SFILES 2.0 repository (\url{https://doi.org/10.5281/zenodo.6901932}) to convert the flowsheets from the SFILES 2.0 format to a NetworkX graph. As described in Section \ref{sec:2023_Graph-to-SFILES_inf_repr_process_graphs}, unit operations, valves, and controllers are represented as nodes and streams and signals as edges.

\begin{figure}[h]
    \centering
    \includegraphics[width=0.45\textwidth]{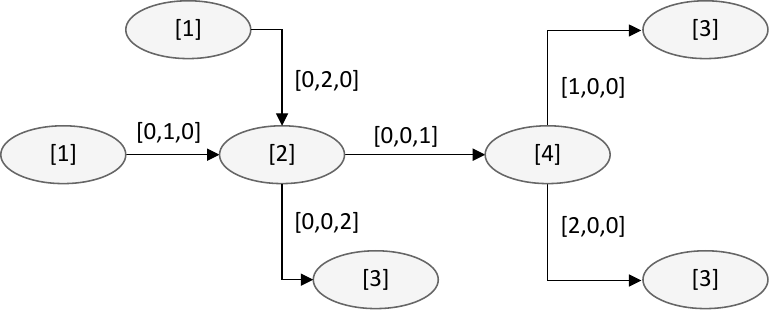}
    \caption{Graph representation of a PFD, Input to the graph encoder.}
    \label{fig:2023_Graph-to-SFILES_encoder_input}
\end{figure}

Secondly, we convert the attributes in the NetworkX graph to node and edge features. The node attributes, specifically unit types, are encoded in a one-dimensional vector using a dictionary. The dictionary stores a relationship between the unit type and an integer value, for instance, \texttt{(raw)} equals 1, \texttt{(hex)} equals 2, etc. The edge attributes are translated into a three-dimensional vector. The first dimension encodes information from which side of a unit a stream is leaving. The second and third dimensions encode to which side of a heat exchanger a stream is connected, according to the scheme shown in Table \ref{tab:Edge_enc}. 
These two pre-processing steps yield the processed graph shown in Figure \ref{fig:2023_Graph-to-SFILES_encoder_input}. 
 
\begin{table}[h]
    \caption{Encoding scheme used for edge attributes.}
    \label{tab:Edge_enc}
    \centering
    \begin{tabular}{llrrr}
        \toprule
        Feature & \multirow{2}{*}{Edge attribute} & \multicolumn{3}{c}{Encoding value} \\
        dimension    & &0      &1      &2 \\
        \midrule
        1& Outlet-stream position   &Unspecified &Top   &Bottom \\
        2& Hex inlet     &Unspecified &Side 1 &Side 2 \\
        3& Hex outlet    &Unspecified &Side 1 &Side 2 \\
        \bottomrule
    \end{tabular}
\end{table}

Similarly to the input data, we pre-process the output data. 
We convert the output SFILES strings to a numerical vector using the tokenizer by \citet{Hirtreiter2023_AutomaticGenerationControl}. The tokenizer breaks the SFILES string into tokens from the SFILES vocabulary. Then the tokens are mapped to a numerical vector which can be processed by the Graph-to-SFILES model.
Lastly, the input graph and output SFILES are combined into a PyTorch Geometric data object.

\section{Results and discussion} \label{sec:2023_Graph-to-SFILES_results}

In the following, we describe and discuss the performance of the Graph-to-SFILES model when trained on a dataset of synthetic PFD-CEF pairs.
In Section \ref{sec:2023_Graph-to-SFILES_results_evaluation}, we report evaluation metrics for the Graph-to-SFILES model and compare the four model architectures. In Section \ref{sec:2023_Graph-to-SFILES_results_comparison}, we compare the Graph-to-SFILES model and our previous sequence-to-sequence approach \citep{Hirtreiter2023_AutomaticGenerationControl}.
In addition, we select an illustrative example from the test dataset and present the model predictions in Section \ref{sec:2023_Graph-to-SFILES_results_example}. Finally, we discuss the results (Section \ref{sec:2023_Graph-to-SFILES_results_discussion}).

\subsection{Graph-to-SFILES evaluation}\label{sec:2023_Graph-to-SFILES_results_evaluation}

We train the Graph-to-SFILES model using four different graph encoder architectures (Section \ref{sec:2023_Graph-to-SFILES_method_encoder}) on the 1k, 10k, and 100k dataset from \citet{Hirtreiter2023_AutomaticGenerationControl}.
All model trainings are performed on a Windows Server 2022 Datacenter with 256GB RAM and an NVIDIA A100 GPU.
We optimize hyperparameters for each architecture using a grid search. 
In total, we run 288 trainings on the 10k dataset for 7.3 days with an average duration of 36.7 minutes per training. We select the optimal hyperparameters based on the validation loss. The optimal hyperparameters and a reference to the full grid search are given in Table \ref{tab:2023_Graph-to-SFILES_hyperparameters}. We observe that the learning rate and the number of decoder layers are the most important hyperparameters as they have the highest correlation with the validation loss.
The learning rate is especially important for the training to converge. Trainings that do not converge have a relatively high learning rate of 1.5e-3 or 2e-3 in common.
In addition, the validation loss is generally lower for larger model architectures. For three out of the four models, the optimal architecture has the maximum number of encoder and decoder layers. 
Furthermore, the effect of the batch size was insignificant.
Then, we apply the optimal hyperparameters to train five model samples for the 1k and 10k datasets and three model samples for the 100k dataset, as well as the corresponding datasets without valves. The model samples have a different weight initialization to gain a representative model performance. Each training runs for 100 episodes. Finally, we evaluate all models on an independent test dataset and summarize the results, including the mean and standard deviation of the samples, in Table \ref{tab:2023_Graph-to-SFILES_results}.

We analyze the Graph-to-SFILES model performance using three evaluation metrics.
First, the models are evaluated on the CEF accuracy. The CEF accuracy measures the models' capabilities of fully recreating the CEF based on a given PFD, equivalent to the accuracy used by \citet{Hirtreiter2023_AutomaticGenerationControl}.
Second, we calculate the PFD reconstruction accuracy.
Due to the autoregressive Graph-to-SFILES decoder, the model is also tasked with recreating the input PFD as an SFILES string. This means that, besides making an error in predicting a control structure, the model can also make an error in generating the underlying process. Hence, to evaluate whether a model also has difficulties generating the underlying process of a flowsheet, we introduce the PFD reconstruction accuracy metric, which only focuses on the process side of the output. We remove the control structures of the predicted CEF, yielding PFD-like schematics. These newly obtained schematics are then compared to the PFD fed to the model. 
Thirdly, we calculate the Bilingual Evaluation Understudy (BLEU) score \citep{Papineni2002_BleuMethodAutomatic}.
Since the CEF accuracy only considers a CEF correct if the predicted SFILES matches the ground truth SFILES exactly, the CEF accuracy cannot differentiate the degree of error. To overcome this, the BLEU score is implemented. BLEU is a scoring metric commonly used in NLP to evaluate the quality of a machine-translated text. The BLEU score divides the predicted SFILES 2.0 into segments and compares each segment to the ground truth segment. The score is then calculated as a percentage and averaged over the entire dataset. Here, we use a segment length of ten.
As explained in Section \ref{sec:2023_Graph-to-SFILES_method_decoding_strategy}, during inference the model utilizes a beam search-based decoding strategy, yielding the top-$k$ predictions sorted by their joined probability. Using these predictions, a top-$k$ score can be calculated for the three evaluation metrics. As multiple valid control structures can exist, it is important to evaluate multiple model outputs. This reflects that the engineer can choose among multiple control options.
Finally, we convert all predicted SFILES into their canonical form before calculating the evaluation metrics.

\begin{table}[h]
\caption{Performance of the Graph-to-SFILES model on an independent test dataset using different encoder architectures and different training dataset sizes. We report the mean and the standard deviation for multiple model samples. The best performance per dataset and evaluation metric is highlighted in bold font.}
\label{tab:2023_Graph-to-SFILES_results}
\begin{tabularx}{\textwidth}{>{\hsize=1.7\hsize}X|>{\hsize=0.6\hsize}X|*{2}{>{\raggedleft\arraybackslash\hsize=0.7\hsize}X}|*{2}{>{\raggedleft\arraybackslash\hsize=0.7\hsize}X}|*{2}{>{\raggedleft\arraybackslash\hsize=0.7\hsize}X}}

\toprule
\textbf{Model architecture} & \textbf{Dataset} & \multicolumn{2}{c}{\textbf{CEF accuracy}} & \multicolumn{2}{c}{\textbf{PFD reconstruction}} & \multicolumn{2}{c}{\textbf{BLEU}}\\ 
\midrule
& & Top-1 (\%) & Top-5 (\%) & Top-1 (\%) & Top-5 (\%) & Top-1 (\%) & Top-5 (\%) \\
\midrule
\multirow{3}{=}{Graph-to-SFILES (GATv2 encoder)}   &1k & 0.5 \textpm 0.3  & 1.2 \textpm 0.7  & 1.0 \textpm 0.3  & 2.2 \textpm 1.1  & 11.8 \textpm 1.6  & 17.3 \textpm 2.2 \\
&10k & 36.0 \textpm 1.9  & 64.9 \textpm 2.0  & 65.8 \textpm 2.1  & 83.1 \textpm 1.6  & 62.5 \textpm 1.0  & 82.0 \textpm 1.3 \\
&100k & 45.1 \textpm 0.8  & 77.4 \textpm 1.6  & 79.2 \textpm 1.0  & 94.7 \textpm 0.8  & 68.6 \textpm 0.4  & 88.6 \textpm 0.6 \\
\midrule
\multirow{3}{=}{Graph-to-SFILES (GraphConv encoder)}      &1k & 5.6 \textpm 1.1  & 10.9 \textpm 1.7  & 9.9 \textpm 1.5  & 17.4 \textpm 2.2  & 31.5 \textpm 1.4  & 41.8 \textpm 1.8 \\
&10k & 39.2 \textpm 2.2  & 68.3 \textpm 2.6  & 69.7 \textpm 4.3  & 86.4 \textpm 2.9  & 64.5 \textpm 1.2  & 83.9 \textpm 1.3 \\
&100k & 45.9 \textpm 0.4  & 78.8 \textpm 0.6  & 80.3 \textpm 1.7  & 96.7 \textpm 0.6  & 69.1 \textpm 0.4  & 89.8 \textpm 0.3 \\
\midrule
\multirow{3}{=}{Graph-to-SFILES (Graph Transformer encoder)}
&1k & 10.5 \textpm 0.7  & 21.9 \textpm 1.5  & 19.1 \textpm 1.5  & 34.2 \textpm 2.0  & 39.1 \textpm 0.6  & 52.1 \textpm 0.5 \\
&10k & 29.0 \textpm 0.2  & 63.5 \textpm 1.2  & 50.9 \textpm 2.1  & 81.9 \textpm 2.6  & 60.5 \textpm 1.2  & 81.7 \textpm 0.7 \\
&100k & 30.0 \textpm 3.0  & 63.0 \textpm 5.6  & 46.3 \textpm 3.6  & 82.1 \textpm 6.2  & 61.1 \textpm 2.4  & 81.4 \textpm 3.6 \\
\midrule
\multirow{3}{=}{Graph-to-SFILES (Combined model encoder)}    &1k & \textbf{15.2 \textpm 0.1}  & \textbf{28.4 \textpm 1.7}  & \textbf{28.5 \textpm 1.5}  & \textbf{41.8 \textpm 1.4}  & \textbf{43.2 \textpm 1.1}  & \textbf{57.7 \textpm 0.8} \\
&10k & \textbf{41.9 \textpm 0.6}  & 73.2 \textpm 0.9  & 74.0 \textpm 0.2  & 90.8 \textpm 1.3  & \textbf{66.7 \textpm 0.6}  & 87.0 \textpm 0.6 \\
&100k & 46.1 \textpm 0.7  & 79.0 \textpm 0.3  & 80.5 \textpm 0.5  & 96.8 \textpm 0.5  & 69.2 \textpm 0.6  & 89.8 \textpm 0.2 \\
\midrule
\midrule
\multirow{3}{=}{Sequence-to-sequence \citep{Hirtreiter2023_AutomaticGenerationControl}}    &1k & 0.3\rule{4.5ex}{0pt}  & 0.9\rule{4.5ex}{0pt}  & 0.3\rule{4.5ex}{0pt}  & 1.7\rule{4.5ex}{0pt}  & 10.5\rule{4.5ex}{0pt}  & 18.7\rule{4.5ex}{0pt} \\
&10k & 37.7\rule{4.5ex}{0pt}  & \textbf{74.8}\rule{4.5ex}{0pt}  & \textbf{81.4}\rule{4.5ex}{0pt}  & \textbf{91.0}\rule{4.5ex}{0pt}  & 63.3\rule{4.5ex}{0pt}  & \textbf{90.1}\rule{4.5ex}{0pt} \\
&100k & \textbf{48.6}\rule{4.5ex}{0pt}  & \textbf{89.2}\rule{4.5ex}{0pt}  & \textbf{99.2}\rule{4.5ex}{0pt}  & \textbf{99.5}\rule{4.5ex}{0pt}  & \textbf{70.8}\rule{4.5ex}{0pt}  & \textbf{96.4}\rule{4.5ex}{0pt} \\
\bottomrule
\end{tabularx}
\end{table}

\begin{table}[h]
\caption{Evaluation on the datasets where the input PFDs do not contain valves (without valves). We report the mean and the standard deviation for multiple model samples. The best performance per evaluation metric on the 10k dataset is highlighted in bold font.}
\label{tab:2023_Graph-to-SFILES_results_without_valves}
\begin{tabularx}{\textwidth}{>{\hsize=1.7\hsize}X|>{\hsize=0.6\hsize}X|*{2}{>{\raggedleft\arraybackslash\hsize=0.7\hsize}X}|*{2}{>{\raggedleft\arraybackslash\hsize=0.7\hsize}X}|*{2}{>{\raggedleft\arraybackslash\hsize=0.7\hsize}X}}

\toprule
\textbf{Model architecture} & \textbf{Dataset} & \multicolumn{2}{c}{\textbf{CEF accuracy}} & \multicolumn{2}{c}{\textbf{PFD reconstruction}} & \multicolumn{2}{c}{\textbf{BLEU}}\\ 
\midrule
& & Top-1 (\%) & Top-5 (\%) & Top-1 (\%) & Top-5 (\%) & Top-1 (\%) & Top-5 (\%) \\
\midrule
\multirow{3}{=}{Graph-to-SFILES (Combined model encoder)}    &1k & 6.7 \textpm 1.2  & 17.3 \textpm 1.3  & 13.4 \textpm 1.7  & 27.2 \textpm 1.4  & 38.0 \textpm 1.2  & 52.6 \textpm 0.6 \\
&10k & \textbf{21.0 \textpm 0.6}  & 46.9 \textpm 1.2  & 37.8 \textpm 0.8  & 63.9 \textpm 0.5  & \textbf{58.9 \textpm 0.4}  & 78.2 \textpm 0.4 \\
&100k & 24.1 \textpm 0.3  & 54.6 \textpm 0.7  & 44.8 \textpm 1.0  & 69.5 \textpm 0.4  & 62.1 \textpm 0.2  & 83.4 \textpm 0.5 \\\midrule
\multirow{3}{=}{Sequence-to-sequence \citep{Hirtreiter2023_AutomaticGenerationControl}} &&&&&&&\\
&10k & 17.8\rule{4.5ex}{0pt}  & \textbf{48.0}\rule{4.5ex}{0pt}  & \textbf{41.5}\rule{4.5ex}{0pt}  & \textbf{67.7}\rule{4.5ex}{0pt}  & 55.8\rule{4.5ex}{0pt}  & \textbf{81.6}\rule{4.5ex}{0pt} \\
&&&&&&&\\
\bottomrule
\end{tabularx}
\end{table}

First, we compare the graph-to-sequence model architectures proposed in this work.
When comparing the CEF accuracy, the results indicate that the Combined model encoder performs best. It reaches the highest CEF accuracy for the 1k (top-1: 15.2\%), 10k (top-1: 41.9\%), and 100k (top-1: 46.1\%) dataset. 
For the 1k dataset, the performance of the other graph encoders in descending order is as follows: Graph Transformer (top-1: 10.5\%), GraphConv (top-1: 5.6\%), and GATv2 (top-1: 0.5\%).
While the GATv2-based model and the GraphConv-based model perform similarly to the Combined model-based model for the 10k and 100k datasets, the GraphTransformer-based model performs significantly worse. 
We observe similar results when comparing the four graph encoders on the PFD reconstruction accuracy and the BLEU score. The Combined model reaches the highest scores for all datasets, both for the top-1 and top-5 scores. 
The GATv2 model scores lowest for the 1k dataset, for instance, a top-1 BLEU score of 11.8\% while the other architectures score over 30\%.
Table \ref{tab:2023_Graph-to-SFILES_results} shows that the GATv2 and the GraphConv reach a similar score to the Combined model for the 10k and 100k datasets. For instance, the GATv2 and GraphConv reach a top-5 PFD reconstruction accuracy for the 100k dataset of 94.7\% and 96.7\% respectively, less than 2.1\% short of the Combined model. 
We conclude that the Combined model-based architecture is best suited for the given chemical process data. 
In the following, we further analyze the Graph-to-SFILES model using the Combined model encoder.

The results suggest that the Graph-to-SFILES model can successfully learn to read process graphs and predict the control structure using the SFILES notation. For instance, 46.1\% of control structures are correctly predicted when we use 100,000 training samples.

We can observe that the model performance significantly increases with the dataset size. A closer inspection of the table indicates that the marginal benefits of enlarging the dataset decrease with the dataset size. While the CEF top-1 accuracy increases by 180\% from the 1k to the 10k dataset, it increases only by 12\% from the 10k to the 100k dataset. We observe a similar trend for the PFD reconstruction accuracy and the BLEU score.

In addition, we compare the CEF and the PFD reconstruction accuracies. The PFD reconstruction is an easier task than the CEF prediction, as the PFD is a subset of the CEF. From the data in Table \ref{tab:2023_Graph-to-SFILES_results}, it is apparent that the model makes mistakes both for the PFD reconstruction and the control structure prediction. The PFD reconstruction accuracy is below 100\% for all datasets: The PFD reconstruction accuracy reaches 28.5\% (1k dataset), 74\% (10k dataset), and 80.5\% (100k dataset). The CEF accuracy is always below the PFD reconstruction accuracy, indicating additional mistakes in the control structure prediction. 

Another interesting trend is the performance on the smallest, 1k dataset. The BLEU score reaches 43.2\% (top-1) and 57.7\% (top-5). This shows that if we divide the model output into segments of ten tokens, the Graph-to-SFILES model can predict around half of the output segments correctly. In addition, even with a relatively small dataset, the top-1 CEF accuracy is 15.2\%.
We expect that the performance especially on the smaller scale dataset, e.g., with 1,000 training flowsheets, is relevant. More detailed and industrially-relevant datasets are likely to be small-scale and require data-efficient model architectures.

\begin{figure}[h]
    \centering
    \begin{subfigure}[c]{0.48\textwidth}
        \centering
        \includegraphics[width=\textwidth]{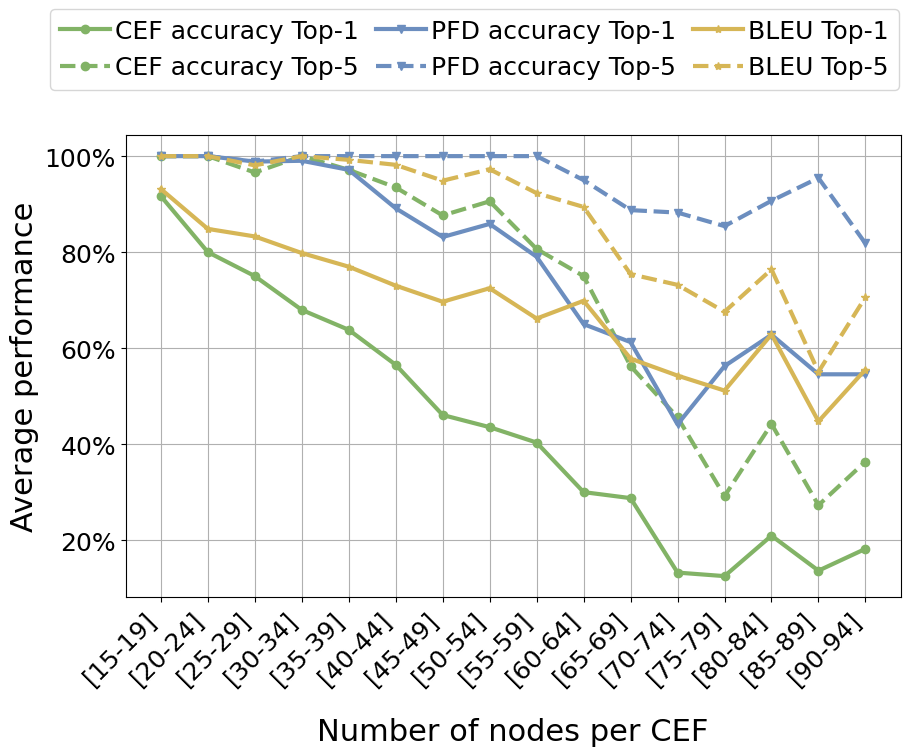}
        \caption{Flowsheet size analysis.}
        \label{fig:Graph-to-SFILES_flowsheet_complexity}
    \end{subfigure}
    \hfill
    \begin{subfigure}[c]{0.48\textwidth}
        \centering
        \includegraphics[width=\textwidth]{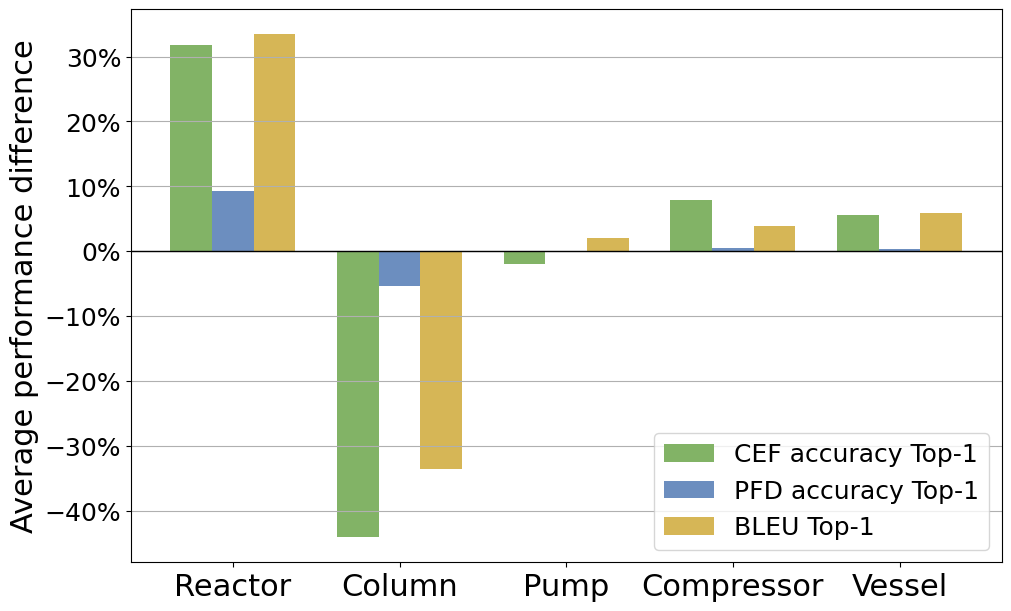}
        \caption{Equipment-specific analysis.}
        \label{fig:Graph-to-SFILES_unit_type_performance}    
    \end{subfigure}
    \caption{Detailed analysis of Graph-to-SFILES (Combined model encoder) performance. Model is trained on 100k dataset. The figure shows (a) the average performance of the model for different ranges of flowsheet sizes and (b) the difference between the average top-1 evaluation metrics for flowsheets where a specific equipment is present or absent.}
    \label{fig:Graph-to-SFILES_detailed_analysis}
\end{figure}

Finally, we perform a detailed analysis of the Combined model encoder trained on the 100k dataset. Figure \ref{fig:Graph-to-SFILES_flowsheet_complexity} shows the average performance of the model for different ranges of flowsheet sizes. We observe that for larger CEFs, all evaluation metrics decrease. For instance, the top-1 CEF accuracy decreases from over 80\% for flowsheets with up to 24 nodes to less than 20\% for flowsheets with more than 85 nodes. Interestingly, the relative error density increases for larger flowsheets, not just the absolute number of incorrect predictions. This is reflected in the decreasing BLEU score with increasing flowsheet size. A possible explanation is the autoregressive nature of the decoder—once an incorrect token is predicted, it can affect subsequent predictions. Further analysis is needed to confirm this. 
Besides the flowsheet size, we analyze the impact of the presence of different equipment on the top-1 performance (Figure \ref{fig:Graph-to-SFILES_unit_type_performance}). We calculate the average top-1 evaluation metrics for flowsheets where the equipment is present or absent. In Figure \ref{fig:Graph-to-SFILES_unit_type_performance}, we report the difference between these values. For the equipment pump, compressor, and vessel, the difference is below 10\% and relatively low. Furthermore, the CEF accuracy and BLEU score increase by more than 30\% if a reactor is present while they decrease by more than 30\% for a distillation column. Considering that the PFD accuracy only changes by less than 10\% in both cases, the performance differences are likely related to the control structure. As mentioned in Section \ref{sec:2023_Graph-to-SFILES_data}, multiple valid control structures can exist for a PFD, resulting in a one-to-many mapping. Especially the distillation column pattern has multiple control options. Therefore, the performance metrics likely underestimate the model performance when distillation columns are present.

\subsection{Comparison Graph-to-SFILES and sequence-to-sequence model}\label{sec:2023_Graph-to-SFILES_results_comparison}

A comparison between the Graph-to-SFILES model using the Combined model encoder and the sequence-to-sequence model by \citet{Hirtreiter2023_AutomaticGenerationControl} suggests that the model performance depends on the dataset size. 
On the 1k dataset, we see significantly better performance when using a graph encoder instead of a sequence encoder. Table \ref{tab:2023_Graph-to-SFILES_results} shows that the CEF accuracy increases from 0.3\% to 15.2\% (top-1) and from 0.9\% to 28.4\% (top-5). 
By contrast, the sequence-to-sequence model reaches higher scores than the Graph-to-SFILES model in all evaluation metrics for the 100k dataset. While the sequence-to-sequence model has a CEF accuracy of 48.6\% (top-1) and 89.2\% (top-5), the Graph-to-SFILES model scores lower with 46.1\% (top-1) and 79\% (top-5). Especially the top-1 PFD reconstruction accuracy is significantly higher for the sequence-to-sequence model at 99.2\% compared to the Graph-to-SFILES model at 80.5\%.
On the 10k dataset, we observe a similar performance between the Graph-to-SFILES and the sequence-to-sequence model. Here, the Graph-to-SFILES model reaches a higher top-1 CEF accuracy and BLEU score and the sequence-to-sequence reaches higher scores in the other evaluation metrics.
Likewise, both models yield similar results when compared to a dataset where the input PFDs do not contain valves (10k without valves) as shown in Table \ref{tab:2023_Graph-to-SFILES_results_without_valves}.

\citet{Hirtreiter2023_AutomaticGenerationControl} also trained the sequence-to-sequence model on a small dataset of 250 P\&IDs from industrial and academic sources and on a small dataset of 100 synthetic flowsheets. They report accuracies of 0\%. We train the Graph-to-SFILES model on the same datasets. First, we use the 250 P\&IDs from industrial and academic sources to fine-tune the Combined model trained on the 100k dataset. However, the model evaluation showed very low scores for all evaluation metrics. We presume that the distribution of the 250 P\&ID graphs from industrial and academic sources is too different from the synthetic dataset used for pre-training. Second, we train the Graph-to-SFILES model on the 100 synthetic flowsheets. The resulting CEF accuracy and PFD reconstruction accuracy are below 1\%, suggesting that a larger training dataset is needed to successfully predict control structures.

\subsection{Illustrative example}\label{sec:2023_Graph-to-SFILES_results_example}

\begin{figure}[h]
    \centering
    \includegraphics[width=0.45\textwidth]{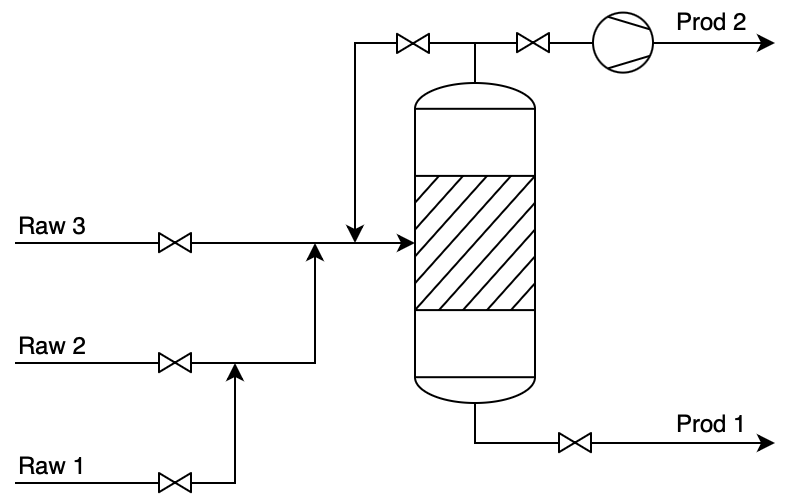}
    \caption{Illustrative example PFD.}
    \label{fig:2023_Graph_to_SFILES_case_study}
\end{figure}

To illustrate the Graph-to-SFILES model performance, we predict the CEF for a sample PFD from the independent test dataset. The flowsheet, shown in Figure \ref{fig:2023_Graph_to_SFILES_case_study}, contains three raw material streams which are mixed and fed to a reactor. The reactor has two outlets of which one is split to be partly recycled and partly compressed.

\begin{figure}[h]
    \centering
    \begin{subfigure}[b]{0.45\textwidth}
        \includegraphics[width=\textwidth]{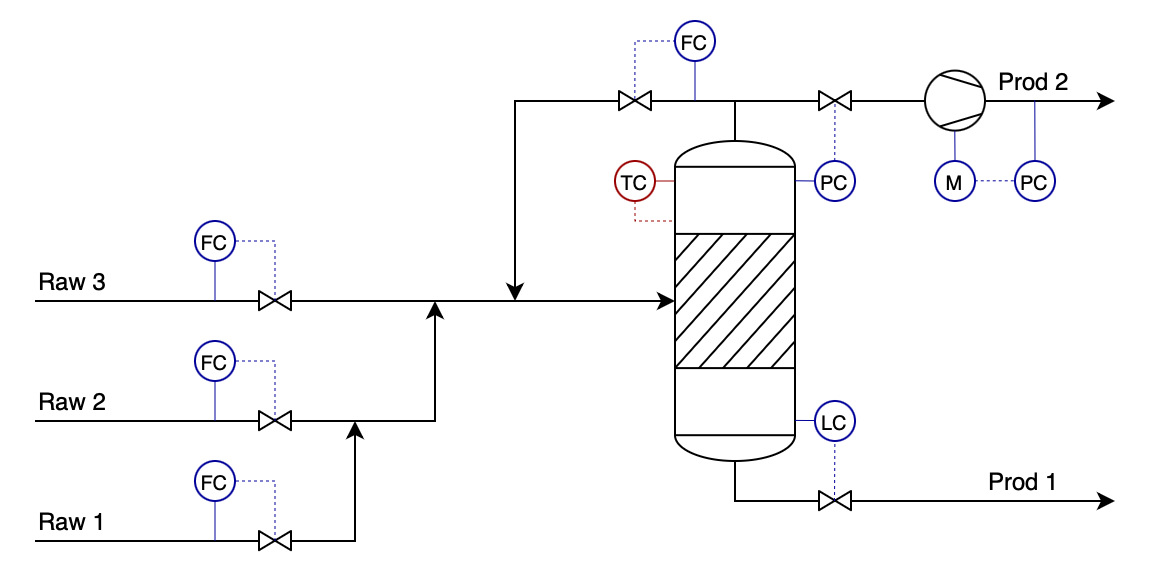}
        \caption{First prediction, incorrect. The model predicts a temperature controller instead of a temperature indicator for the reactor.}
        \label{fig:2023_Graph_to_SFILES_prediction_1}
    \end{subfigure}
    \hfill
    \begin{subfigure}[b]{0.45\textwidth}
        \includegraphics[width=\textwidth]{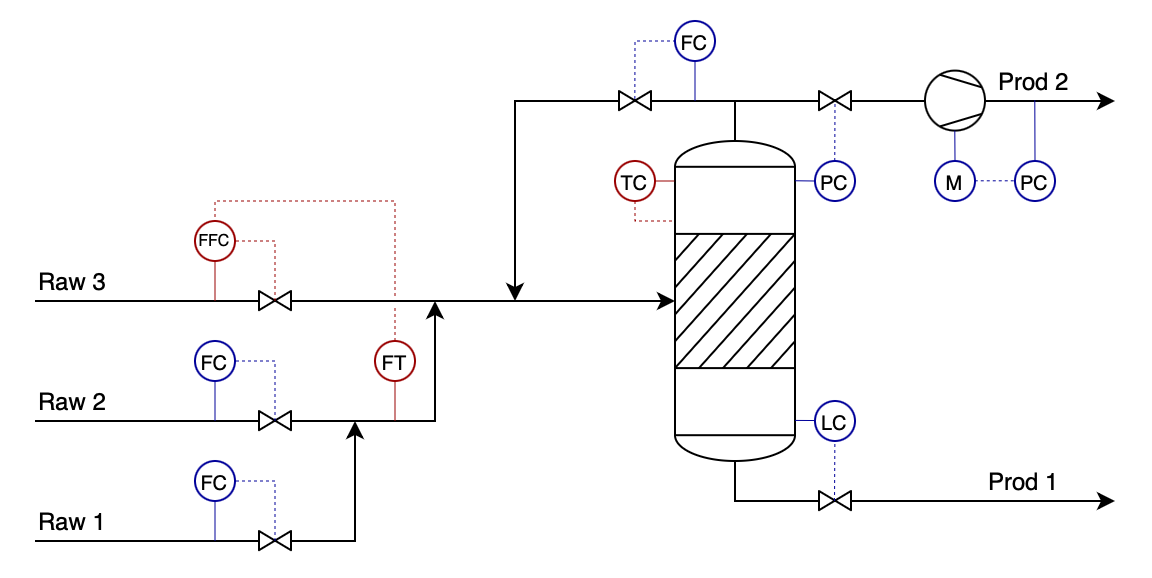}
        \caption{Second prediction, incorrect. Raw 3 should be controlled with a regular flow controller. In addition, the model predicts a temperature controller instead of a temperature indicator for the reactor.}
        \label{fig:2023_Graph_to_SFILES_prediction_2}
    \end{subfigure}
    \hfill
    \begin{subfigure}[b]{0.45\textwidth}
        \includegraphics[width=\textwidth]{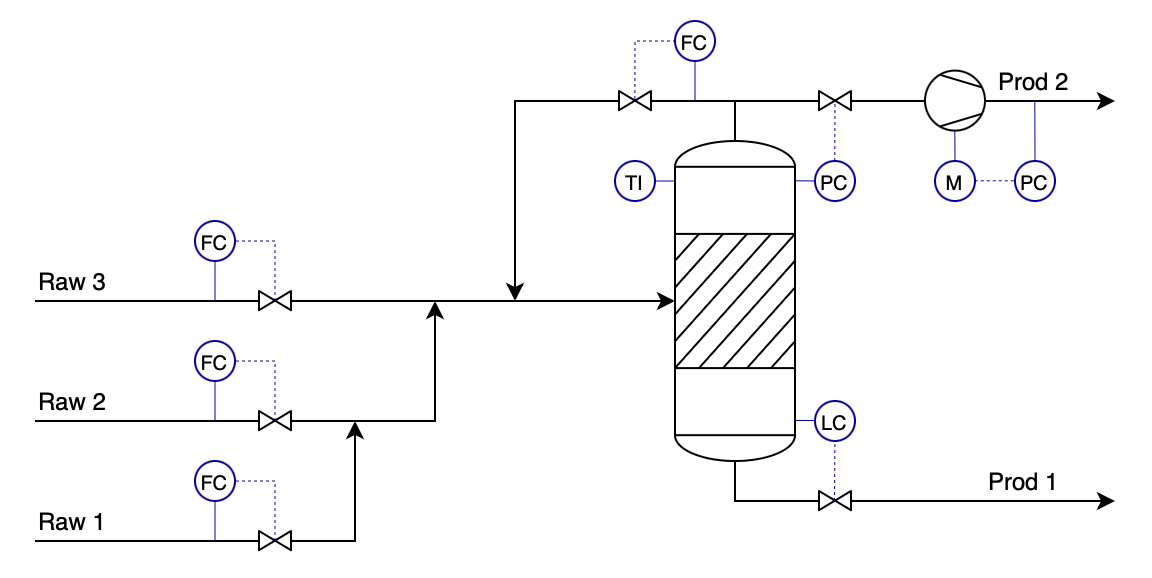}
        \caption{Third prediction, correct.}
        \label{fig:2023_Graph_to_SFILES_prediction_3}
    \end{subfigure}
    \hfill
    \begin{subfigure}[b]{0.45\textwidth}
        \includegraphics[width=\textwidth]{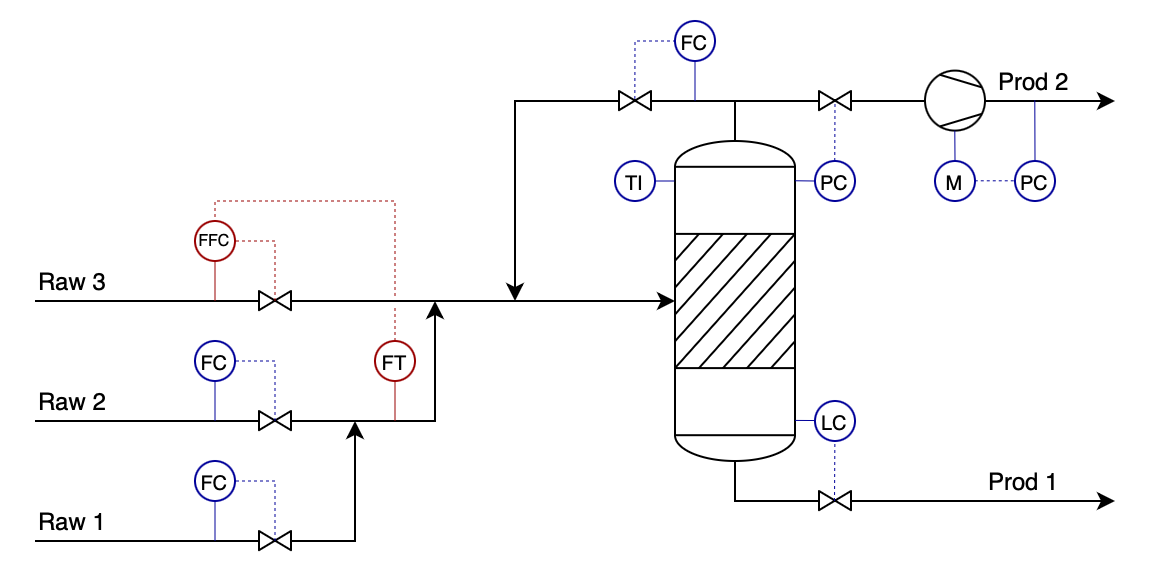}
        \caption{Fourth prediction, incorrect. Raw 3 should be controlled with a regular flow controller.}
        \label{fig:2023_Graph_to_SFILES_prediction_4}
    \end{subfigure}
    \hfill
    \begin{subfigure}[b]{0.45\textwidth}
        \includegraphics[width=\textwidth]{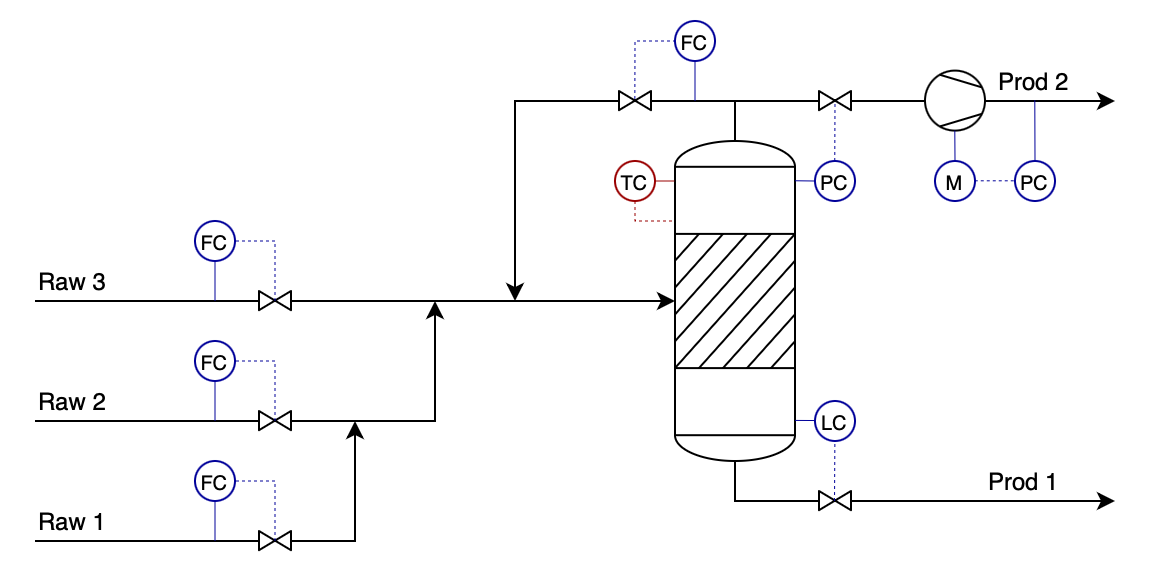}
        \caption{Fifth prediction, incorrect. The model predicts a temperature controller instead of a temperature indicator for the reactor.}
        \label{fig:2023_Graph_to_SFILES_prediction_5}
    \end{subfigure}
    \caption{Control structure prediction for the sample PFD (colored black) where correct predictions are shown in blue while incorrect predictions are shown in red. The five predictions correspond to the five beams with the highest joined probability during beam search decoding, ordered by descending joint probability.}
    \label{fig:2023_Graph_to_SFILES_prediction}
\end{figure}

We employ the Combined model encoder trained on 100,000 flowsheets. From the three model samples presented in \ref{tab:2023_Graph-to-SFILES_results}, we select the sample with the highest top-1 CEF accuracy. The model predicts the following five SFILES 2.0. Here, the five predictions correspond to the five beams with the highest joined probability during beam search decoding, ordered by descending joint probability.
Figure \ref{fig:2023_Graph_to_SFILES_prediction} illustrates the predictions as flowsheets, in which the correctly added controllers are shown in blue, whereas incorrect additions are marked in red.

\begin{enumerate}[itemsep=10pt]
\item \begin{verbatim}
(raw)(C){FC}_1(v)<_1(mix)<&|(raw)(C){FC}_2(v)&<_2|(mix)<&|(raw)(C){FC}_3(v)&<_3|
(mix)<1(r)<_4[(C){TC}_4][(C){LC}_5][{bout}(v)<_5(prod)]{tout}(C){PC}_6(v)<_6(splt)
[(comp)[(C){M}<_7](C){PC}_7(prod)](C){FC}_8(v)1<_8
\end{verbatim}

\item \begin{verbatim}
(raw)(C){FC}_1(v)<_1(mix)<&|(raw)(C){FC}_2(v)&<_2|(C){FT}_3(mix)<&|(raw)(C){FFC}
_4<_3(v)&<_4|(mix)<1(r)<_5[(C){TC}_5][(C){LC}_6][{bout}(v)<_6(prod)]{tout}(C){PC}
_7(v)<_7(splt)[(comp)[(C){M}<_8](C){PC}_8(prod)](C){FC}_9(v)1<_9
\end{verbatim}

\item \begin{verbatim}
(raw)(C){FC}_1(v)<_1(mix)<&|(raw)(C){FC}_2(v)&<_2|(mix)<&|(raw)(C){FC}_3(v)&<_3|
(mix)<1(r)[(C){TI}][(C){LC}_4][{bout}(v)<_4(prod)]{tout}(C){PC}_5(v)<_5(splt)
[(comp)[(C){M}<_6](C){PC}_6(prod)](C){FC}_7(v)1<_7
\end{verbatim}

\item \begin{verbatim}
(raw)(C){FC}_1(v)<_1(mix)<&|(raw)(C){FC}_2(v)&<_2|(C){FT}_3(mix)<&|(raw)(C){FFC}
_4<_3(v)&<_4|(mix)<1(r)[(C){TI}][(C){LC}_5][{bout}(v)<_5(prod)]{tout}(C){PC}_6
(v)<_6(splt)[(comp)[(C){M}<_7](C){PC}_7(prod)](C){FC}_8(v)1<_8
\end{verbatim}

\item \begin{verbatim}
(raw)(C){FC}_1(v)<_1(mix)<&|(raw)(C){FC}_2(v)&<_2|(mix)<&|(raw)(C){FC}_3(v)&<_3|
(mix)<1(r)<_4[(C){TC}_4][(C){LC}_5][{bout}(v)<_5(prod)]{tout}(C){PC}_6(v)<_6(splt)
[(C){FC}_7(v)1<_7](comp)[(C){M}<_8](C){PC}_8(prod)
\end{verbatim}
\end{enumerate}

We compare the model predictions with the ground truth CEF. The third prediction (Figure \ref{fig:2023_Graph_to_SFILES_prediction_3}) is equal to the ground truth CEF. Each raw material stream is controlled independently by a flow controller. The reactor has a temperature indicator, a level controller to control the bottom outlet, and a pressure controller to control the top outlet. In addition, the recycle contains a flow controller. Finally, a pressure controller acts on a motor to control the compressor at the top outlet.
When looking at the model prediction above, we identify the following trends:

\begin{itemize}
\item[(i)] The syntax of all five generated SFILES is correct. 

\item[(ii)] The model correctly regenerates the PFD in all five cases.

\item[(iii)] The model predicts the majority of controllers correctly. In particular, the third prediction is correct and the top 5 predictions have at least 6 out of 8 controllers correctly predicted. This indicates that the model learns to read the process graph and to predict a suitable control structure.

\item[(iv)] The first and fifth predictions are identical in their flowsheet format, but in the SFILES notation they are not. The first prediction is the canonical SFILES which first adds the outlet stream and then closes the recycle. In contrast, the recycle and outlet order is reversed in the fifth prediction.
This shows that the model does not always generate canonical, or unique, SFILES representations.

\item[(v)] We observe that the model makes two types of predictions that differ from the ground truth. 
First, it predicts a temperature controller (TC) instead of an indicator (TI) for the reactor in the first, second, and fifth predictions. 
A temperature indicator would be sufficient for an adiabatic reactor, while a temperature controller is required for a diabatic reactor. Here, it is not known whether the reactor is adiabatic or diabatic.
Second, it predicts a flow ratio controller (FFC) for the inlet Raw 3 instead of a regular flow controller (FC) in the second and fourth predictions. The flow ratio controller keeps the reactants in the desired proportion instead of controlling the flow of the reactants individually.
In both cases, the two control alternatives are valid process configurations and are present in the training data.
A possible explanation for the wrong predictions could be limited information in the model input. The PFD data do not contain any process-specific information. Thus, the model has no information to determine the correct control structure other than any potential bias in the dataset. 

\end{itemize}

\subsection{Discussion}\label{sec:2023_Graph-to-SFILES_results_discussion}

Overall, the results indicate that the Graph-to-SFILES model successfully learns node and edge features in process graphs to generate a CEF. 
The Graph-to-SFILES model thereby serves as a proof-of-concept for leveraging graph data in control structure prediction. 
As such, the Graph-to-SFILES model is meant as a support tool for engineers and not to fully automate the control structure design.

A key observation is the difference in performance between the graph encoder and the sequence encoder.
We expect that the graph encoder is more data efficient than the sequence transformer but that for larger-scale data, the difference between the encoders diminishes as the sequence encoder has sufficient data to learn the SFILES 2.0 notation. 
This aligns with our results on the 1k dataset, where the graph encoder outperforms the sequence encoder.
However, our results also show that the sequence encoder performs considerably better for the large-scale, 100k dataset. 
The most likely explanations for the sequence encoder outperforming the graph encoder on the large-scale dataset are:
First, the input and output formats of the sequence-to-sequence model are SFILES. Using the same format in input and output might make it easier for the sequence-to-sequence model to preserve information and reconstruct the PFD topology in the output. We see that especially the PFD reconstruction is significantly better for the sequence-based model compared to the graph-based model.
Second, GNNs are prone to over-smoothing \citep{Rusch2023_SurveyOversmoothingGraph}. For an increasing number of layers in the graph encoder, node features become more similar, up to a point where unique features of a node are lost. This limits the depth of GNNs and therefore the learning capacity.
Related to this, capturing long-range dependencies is challenging for GNNs \citep{Topping2021_UnderstandingOverSquashing,Rusch2023_SurveyOversmoothingGraph}. Contrarily, an attention-based sequence encoder assumes that every token attends to every other token, thereby capturing long-range dependencies efficiently.
While these points are possible explanations, it is subject to ongoing research as to why the sequence-to-sequence model outperforms the Graph-to-SFILES model on large-scale data.

To improve the prediction accuracy of the Graph-to-SFILES model, multiple promising research avenues exist. First, the PFD reconstruction accuracy could be improved as the PFD topology is explicitly known from the model input and should be preserved in the output. We expect that enforcing the PFD topology in the output SFILES will improve the model's performance. For instance, constrained beam search could restrict the predicted unit operations to those present in the input data.
Second, multimodel fusion could improve the model performance \citep{Gao2020_SurveyDeepLearning}. Multimodal fusion aims to integrate data from different modalities. For deep learning, multimodal models have shown superior performance to unimodal models \citep{Xu2023_MultimodalLearningTransformers}. Specifically, one could build an architecture that has the same decoder as the Graph-to-SFILES model but a new, multimodal encoder that leverages both SFILES and graph representation.
Third, many control structures depend on the field and safety regulations. Including field and safety regulations could be achieved through: (i) domain-specific attributes in node and edge features, (ii) fine-tuning on field-specific datasets, and (iii) enforcing safety constraints via expert systems in a hybrid AI approach.

Furthermore, it is important to bear in mind the limitations of the current work.
Firstly, the current dataset does not include any equipment, operation, or stream parameters but solely topological information. The proof needs to be made that the Graph-to-SFILES model can learn those parameters and that they improve the model prediction.
Secondly, we train our model on a synthetic dataset. In an ideal scenario, we would train our model on curated, industrial process data available at a large scale. Further efforts need to be made to collect data that are close to the final application in industrial practice, e.g., by digitizing industry data \citep{Theisen2023_DigitizationChemicalProcess} or by mining academic data and patents \citep{Schweidtmann2024_MiningChemicalProcess,Balhorn2022_FlowsheetRecognitionUsing}.
Thirdly, our Graph-to-SFILES model is a purely data-driven approach. Data-driven approaches proved to be successful for generative AI tasks if large-scale data are available, for instance for molecular design \citep{Du2024_MachineLearningAided} or language generation (ChatGPT). However, as described above, a large-scale process dataset is currently missing. Therefore, including physical knowledge in our model could potentially improve its performance. Furthermore, physical knowledge could increase safety, interpretability, and trust in the model. Process modeling and control principles could be integrated into the model as hybrid AI approaches.
In consequence, we believe that overcoming the aforementioned limitations brings the vision of AI-assisted control structure design within reach.

Another important aspect to consider in control structure design is that often multiple valid solutions exist for a given process. 
Our dataset accounts for this by including multiple control options for the same PFD.
However, this leads to a one-to-many mapping in the dataset. 
Especially if only the topology of the process is considered, selecting a specific control structure can lead to ambiguous results.
However, the top-1 CEF accuracy only counts a prediction as correct if the predicted CEF matches the ground-truth CEF exactly. Therefore, there is a theoretical maximum top-1 CEF accuracy that lies well below 100\% given the one-to-many mapping.
This can be observed as the top-1 CEF accuracy only marginally improves as we increase the dataset from 10,000 to 100,000 samples. Hence, it is important to also calculate the top-k accuracies.
We acknowledge that adding further process details such as process dynamics and safety constraints could make the control structure prediction more specific.
In general, using a probabilistic model such as the transformer-based Graph-to-SFILES model is advantageous. The Graph-to-SFILES model outputs a probability distribution. This allows to obtain multiple control structure predictions with their respective probability. Future work could exploit this by analyzing the prediction uncertainty of the model \citep{Ott2018_AnalyzingUncertaintyNeural}.

More broadly, a key challenge in evaluating AI-driven control structure design is the lack of standardized metrics. 
The current evaluation metrics (i.e., CEF accuracy, PFD accuracy, BLEU) do not adequately capture the quality of CEFs.
A direct comparison between CEFs is difficult because different control structures are designed for different process objectives, work under different constraints, and have varying levels of centralization. Even if two control structures achieve the same setpoint tracking accuracy, they might differ in robustness, adaptability, explainability, or computational cost. 
It remains an open research question how the quality of CEFs, not to mention P\&IDs, can be quantitatively compared.

In addition, the effectiveness of AI for control structure design compared to expert systems remains an open research question. 
A meaningful comparison could be conducted on a set of representative processes. For each process, control structures would be generated using (i) the Graph-to-SFILES model and (ii) an expert system, operated by a trained user. The resulting control structures could then be evaluated based on the following criteria: Correctness, computational efficiency, explainability, and ease of use.
A potential limitation of this comparison is the dependence on the user’s familiarity with the expert system. Since expert systems often require manual rule encoding and extensive training, the quality of results may vary based on the user’s expertise. This factor should be considered when interpreting the comparison results.

\section{Conclusions}\label{sec:2023_Graph-to-SFILES_conclusions}

This work proposes the Graph-to-SFILES model. The Graph-to-SFILES model automatically extends a process schematic with control structures using a graphical representation of the process. 
We propose a GNN architecture for the graph encoder and compare it with three architectures from the literature on a synthetic dataset.

The best-performing Graph-to-SFILES model, using a graph encoder with positional encoding, can predict the correct control structure for 46.1\% of the synthetic test data. Even in cases where the model did not fully predict the correct schematics, the output tends to be a valid alternative control layout, as demonstrated in an illustrative example. However, it remains an open research question whether the Graph-to-SFILES model can learn from industry-relevant and complex process details, such as stream conditions or composition, provided as node and edge features.

Future work will tackle the remaining limitations of our approach. A first step could be to acquire larger datasets from industrial and academic sources with detailed process information. We expect that including more industrially relevant data in the training can greatly improve the usefulness and acceptance of this model in process engineering.
Building on the current work, the Graph-to-SFILES model can also be used for other generative tasks in process design, for instance, the auto-correction of flowsheets \citep{Balhorn2024_AutocorrectionChemicalProcess}, alternative flowsheet suggestion, or HAZOP report generation.


\bibliography{references}

\appendix

\section{Appendix}

\subsection{Combined model}\label{sec:2023_Graph-to-SFILES_appendix_model}

First, the Combined model calculates a positional encoding $\bm{\lambda}_{i}$ for every node $i$ from the graph's normalized Laplacian matrix $\bm{L}_{\rm norm}\in \mathbb{R}^{N \times N}$, where $N$ is the number of nodes in the graph. 

\begin{equation}
    \bm{L}_{\rm norm} = \bm{D}^{-\frac{1}{2}}\bm{LD}^{-\frac{1}{2}}
\end{equation}

Here, $\bm{D}\in \mathbb{R}^{N \times N}$ is the degree matrix, $L=D-A$ the Laplacian matrix, and $\bm{A}\in \mathbb{R}^{N \times N}$ is the adjacency matrix.
The $k$ smallest non-trivial eigenvectors $\bm{\lambda}_i\in\mathbb{R}^{k}$ of $\bm{L}_{\rm norm}$ are added to the original node features $\bm{h}_i^0\in \mathbb{R}^{d_{\rm node}}$ for each node $i$. Both vectors are projected to the encoder embedding dimension $d_{\rm enc}$, with $\bm{W}_{\rm node}\in\mathbb{R}^{d_{\rm enc}\times d_{\rm node}}$ and $\bm{W}_{\lambda}\in\mathbb{R}^{d_{\rm enc}\times k}$, and a bias $\bm{b}\in\mathbb{R}^{d_{\rm enc}}$ is added.

\begin{equation}
    \bm{h}_i = \bm{W}_{\rm node}\bm{h}_i^0 + \bm{W}_{\lambda}\bm{\lambda}_i+\bm{b}_{\rm node} 
\end{equation}

In addition, the edge feature is projected to the embedding dimension:

\begin{equation}
    \bm{e}_{i,j} = \bm{W}_{\rm edge}\bm{e}_{i,j}^0+\bm{b}_{\rm edge},
\end{equation}

with $\bm{W}_{\rm edge} \in\mathbb{R}^{d_{\rm enc}\times d_{\rm edge}}$ and $\bm{e}_{i,j}^0 \in\mathbb{R}^{d_{\rm edge}}$ ($j$ is a neighboring node of node $i$). Note that the Laplacian embedding is only added in the input layer. 
The attention-based message passing block $\alpha_{i,j}^m\in\mathbb{R}$ is calculated for $m$ heads using

\begin{equation}
\alpha_{i,j}^{m} = \text{softmax}\left({\bm{a}^{m}}^{\top} \text{LeakyReLU}\left(\left[\bm{W}_{q_{\rm enc}}^{m}\bm{h}_{i}\Vert\bm{W}_{k_{\rm enc}}^{m}\bm{h}_{j}\Vert\bm{W}_{e_{\rm enc}}^{m}\bm{e}_{i,j}\right]+b^m\right)\right).
\end{equation}

Here, $\bm{h}_i,\bm{h}_j\in\mathbb{R}^{d_{\rm enc}}$ and $\bm{e}\in\mathbb{R}^{d_{\rm enc}}$ are the node and edge features of the previous layer, multiplied by weight matrices $\bm{W}_{q_{\rm enc}}^m,\bm{W}_{k_{\rm enc}}^m,\bm{W}_{e_{\rm enc}}^m\in\mathbb{R}^{d_{\rm enc}^{'}\times d_{\rm enc}}$, concatenated (denoted as $\Vert$), and then summed with a bias $\bm{b}^m\in\mathbb{R}^{3d_{\rm enc}^{'}}$. After applying the nonlinear function LeakyReLU, the output is multiplied by another weight matrix $\bm{a}^m\in\mathbb{R}^{3d_{\rm enc}^{'}}$ and normalized using the softmax function. Here, $d_{\rm enc}$ is the internal encoder dimension and $d_{\rm enc}^{'}=d_{\rm enc}/m$ the head-specific dimension. We make us of the PyTorch Geometric implementation of the GATv2 attention mechanism\footnote{\url{https://pytorch-geometric.readthedocs.io/en/latest/generated/torch_geometric.nn.conv.GATv2Conv.html}}. The node update equation includes (i) message passing from the node's neighbors and (ii) attention head concatenation:

\begin{equation}
\bm{\hat{h}}_i = \concat_{m=1}^M\left( \sum_{j \in \mathcal{N}_{i}}{\text{softmax}(\alpha_{i,j}^m) \bm{W}_{v_{\rm enc}}^m  \bm{h}_j}\right). 
\end{equation}

Finally, the node update is passed through two residual connections with layer normalization and an FFN. Here, the weight matrices have the dimension $\bm{W}_1,\bm{W}_2\in\mathbb{R}^{d_{\rm enc}\times d_{\rm enc}}$. 

\begin{equation}
    \bm{\hat{\hat{h}}}_i=\text{Norm}\left(\bm{h}_i+\bm{\hat{h}}_i\right)
\end{equation}

\begin{equation}
    \bm{\hat{\hat{\hat{h}}}}_i=\bm{W}_{2}\text{ReLU}\left(\bm{W}_{1}\bm{\hat{\hat{h}}}_i+b\right),
\end{equation}
\begin{equation}
    \bm{h}^{'}_i=\text{Norm}\left(\bm{\hat{\hat{h}}}_i+\bm{\hat{\hat{\hat{h}}}}_i\right)
\end{equation}

\subsection{Hyperparameters}

\begin{table}[h]
    \centering
    \caption{Graph-to-SFILES hyperparameters. The complete grid search of hyperparameters can be found here: \url{https://zenodo.org/records/14840560?preview=1&token=eyJhbGciOiJIUzUxMiJ9.eyJpZCI6ImRkYWEyYTMyLTE4M2YtNDZlMC05NGZmLTNlNGYwMmQ5NzVjYiIsImRhdGEiOnt9LCJyYW5kb20iOiIzZmYxODI3NjliOTE2NjQwOWE5M2IzYmYxNTg1ZDM0YyJ9.Zu_uC9F378UUKfHtycMSzJ_qHjv_97r022Otot4EpSLqA-AscjlIcillR_oXsXpP511hqs-VbiHbP8udY9Oaag}.}
    \begin{tabular}{l|r|rrrr}
        \toprule
        Hyperparameter & Values & \multicolumn{4}{c}{Grid search results per model architecture}\\
        \midrule
        & & GATv2 & GraphConv & Graph Transformer & Combined model\\
        \midrule
        Batch size & \{64, 128\} & 128 & 64 & 64 & 64\\
        Learning rate & \{5e-4, 1e-3, 1.5e-3, 2e-3\} & 2e-3 & 1e-3 & 1e-3 & 1e-3\\
        Encoder layers & \{2, 4, 6\} & 4 & 6 & 6 & 6\\
        Decoder layers & \{2, 4, 6\} & 4 & 6 & 6 & 6\\
        Heads $m$ & 8 & - & - & - & -\\
        Embedding $d_{\rm enc}, d_{\rm dec}$ & 128 & - & - & - & -\\
        Dropout & 0.1 & - & - & - & -\\
        Decoder FFN dimension & 2048 & - & - & - & -\\
        \bottomrule
    \end{tabular}
    \label{tab:2023_Graph-to-SFILES_hyperparameters}
\end{table}

\end{document}